\pdfoutput=1

\documentclass{article}

\usepackage[T1]{fontenc}
\usepackage{microtype}
\usepackage{fontawesome5}
\usepackage{graphicx}
\usepackage{subcaption}
\usepackage{booktabs}
\usepackage{array}
\usepackage{placeins}
\usepackage{multirow}
\usepackage{enumitem}
\usepackage{xcolor}
\usepackage{tikz}
\usetikzlibrary{arrows.meta, positioning, calc, shapes.geometric, fit}
\usepackage{pgfplots}
\pgfplotsset{compat=1.18}

\PassOptionsToPackage{hyphens}{url}
\usepackage{hyperref}

\usepackage[preprint]{icml2026}
\ifdefined\ispreprint
  \hypersetup{pdfsubject={}}
\fi

\usepackage[scaled=0.88]{inconsolata}
\usepackage[scaled=0.86,type1,sb]{inter}

\usepackage{amsmath}
\usepackage{amssymb}
\usepackage{mathtools}
\usepackage{amsthm}
\usepackage[capitalize,noabbrev]{cleveref}

\theoremstyle{plain}

\newtheorem{proposition}{Proposition}[section]

\theoremstyle{definition}

\theoremstyle{remark}

\crefname{proposition}{Proposition}{Propositions}
\Crefname{proposition}{Proposition}{Propositions}

\newcommand{\method}{\textsc{ClonAct}}
\newcommand{\mono}{DSH-Mono}
\newcommand{\subneutral}{DSH-Subagent-Neutral}
\newcommand{\subfork}{DSH-Subagent-Fork}
\newcommand{\agentteam}{DSH-AgentTeam}
\newcommand{\superagent}{DSH-SuperAgent}

\definecolor{headerblue}{HTML}{1A3688}
\newcommand{\headerextlink}[1]{\href{#1}{\textcolor{headerblue}{\expandafter\detokenize\expandafter{#1}}}}
\newcommand{\codeurl}{https://github.com/Viacol0916/ClonAct}
\newcommand{\blogurl}{https://viacol.pages.dev/blog/context-as-a-resource/}

\icmltitlerunning{Context as a Resource}

\begin{document}

\twocolumn[
  \icmltitle{Context as a Resource}

  \begin{icmlauthorlist}
    \icmlauthor{Xufei Tian}{ecust}
    \icmlauthor{Han Hu}{ecust}
    \icmlauthor{Wenli Du}{ecust}
  \end{icmlauthorlist}

  \icmlaffiliation{ecust}{East China University of Science and Technology, Shanghai, China}
  \icmlcorrespondingauthor{Wenli Du}{wldu@ecust.edu.cn}

  \ificmlshowauthors
    \vskip 0.10in
    \begin{center}
      {\small
        \raisebox{-0.12ex}{\scalebox{1.05}{\faGithub}}\hspace{0.35em}Code:~\headerextlink{\codeurl}
        \hspace{1.6em}
        \raisebox{-0.12ex}{\scalebox{0.98}{\faGlobe}}\hspace{0.35em}Blog:~\headerextlink{\blogurl}%
      }
    \end{center}
    \vskip -0.06in
  \fi

  \icmlkeywords{LLM Agents, Context Scheduling, Personal Agents, Multi-Agent Systems, Prefix KV-Cache}

  \vskip 0.3in
]

\printAffiliationsAndNotice{}

\begin{abstract}
Managing concurrent tasks across separate agent sessions quickly exhausts human attention, motivating always-on personal agents behind a single continuous interface. Yet when every context shares identical frozen backbone weights, funneling interleaved dialogues and external events into one linear session induces cross-topic constraint bleed and head-of-line stalling. Conversely, router--ticket hierarchies incur lossy prose handoffs, heavy routing overhead, and blind merges---collapsing true-conflict repository merges to a $33.3\%$ three-run mean despite high isolated branch passes. Rather than scheduling anthropomorphic roles, we introduce \method{}, a routerless architecture that virtualizes context blocks directly through a pluggable single-pass non-generative Context MMU, in-place Context Cells, a footprint-aware Synaptic Bus with an isolated trial merge gate, and a wave-flushed Shared Trunk. Across three programmatic multi-stream benchmarks and three model families, our portable trial merge gate lifts true-conflict merges to $63.3\%$, while \method{}'s routerless virtualization attains this ceiling with $45.9\%$ fewer prompt tokens and prevents compact-model router collapse. Under continuous surge streams, \method{} matches top completion ($95.0\%$) with the lowest median wait and zero timed-out status checks.
\end{abstract}

\section{Introduction}
\label{sec:intro}

\begin{figure}[t]
\centering
\definecolor{clonblue}{HTML}{1B6EC2}
\definecolor{trunkamber}{HTML}{B86B00}
\definecolor{spineteal}{HTML}{0F766E}
\definecolor{synred}{HTML}{C0392B}
\definecolor{csuper}{HTML}{E67E22}
\definecolor{slateink}{HTML}{1E293B}
\definecolor{slatemuted}{HTML}{64748B}

\includegraphics{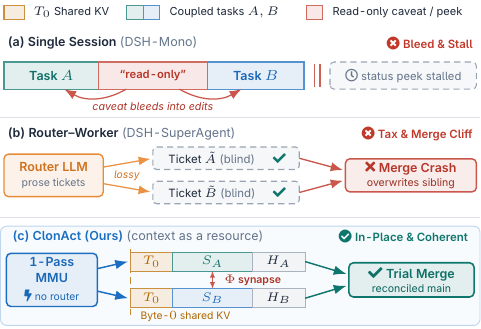}
\vspace{-0.18cm}
\caption{\textbf{Scheduling context blocks directly as virtualized resources.} \textbf{(a)} Flat single-session execution bleeds interleaved caveats across tasks and stalls mid-flight status peeks. \textbf{(b)} Router--worker hierarchies incur generative routing overhead and collapse when blind worktrees merge overlapping edits. \textbf{(c)} \method{} removes the generative router, scheduling byte-$0$ KV-aligned Context Cells in-place with footprint-aware reconciliation.}
\label{fig:teaser}
\vspace{-0.16cm}
\end{figure}

Modern tool-using language model agents can complete a range of coding and workflow tasks end-to-end within an isolated session \citep{yao2023react,yang2024sweagent,wang2025openhands}. In daily work, however, tasks rarely run one at a time: users advance multiple tasks in parallel while asynchronous requests and notifications continue arriving from external channels \citep{froger2026gaia2}. Today, users typically isolate concurrent work by manually opening separate sessions and workspaces, yet monitoring progress, switching focus, and ferrying intermediate findings across many live sessions quickly exhausts human attention \citep{deepseek2026superagentblog,deepseek2026superagentrepo}. Building an always-on personal agent therefore requires hiding session management inside the runtime so that users interact through a single continuous interface \citep{packer2023memgpt,deepseek2026superagentrepo}---raising a central architectural question: \emph{how should an agent system coordinate concurrent tasks behind one unified window when every underlying instance shares the same model weights?}

\begin{figure*}[t]
\centering
\definecolor{cClon}{HTML}{1B6EC2}
\definecolor{cSuper}{HTML}{E67E22}
\definecolor{cMono}{HTML}{3B4252}
\definecolor{cSlateA}{HTML}{7A9BB8}
\definecolor{cSlateB}{HTML}{9AA5B1}
\definecolor{cAlert}{HTML}{C0392B}
\definecolor{slateink}{HTML}{1E293B}

\includegraphics{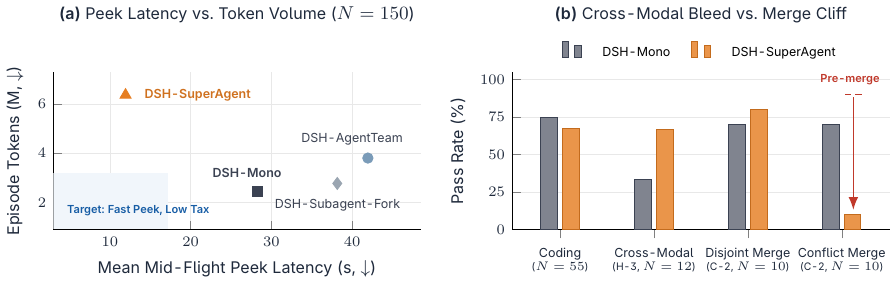}
\vspace{-0.18cm}
\caption{\textbf{Controlled diagnosis of existing paradigms under DSH kernel parity} (Bench-150 Run~1; three-run stability in \cref{tab:c2_mechanistic}b). \textbf{(a)} Single-session and parent-loop designs stall mid-flight status peeks, whereas generative routing unblocks peeks at higher token volume. \textbf{(b)} Flat context degrades under cross-modal caveats, while blind isolated worktrees pass local branch tests yet collapse upon merging overlapping symbols.}
\label{fig:diagnosis}
\vspace{-0.12cm}
\end{figure*}

Long-context studies document sensitivity to input length and evidence position \citep{liu2024lost,levy2024sametask}. In our interleaved-task diagnosis (\cref{sec:diagnosis}), a single linear context (\mono{}, \cref{fig:teaser}a) also exhibits cross-task constraint bleed---a read-only lookup caveat can suppress an in-flight code change---while long tool turns stall status queries by $3.9\times$ (\cref{fig:diagnosis}a). Moreover, as cumulative-request experiments by \citet{deepseek2026superagentblog} demonstrate, subagent delegation in a monolithic session is prompt-dependent and not guaranteed, motivating structural separation behind one entry point via multi-role teams (\agentteam{}; \citealp{hong2024metagpt,wu2024autogen,qian2024chatdev}) or a resident top-level router that summarizes dialogues into isolated worker mandates (\superagent{}, \cref{fig:teaser}b; \citealp{deepseek2026superagentblog,deepseek2026superagentrepo}). Yet unlike human organizations whose heterogeneous expertise requires role division and verbal handoffs, agent instances sharing identical System-2 weights ($\theta_1 = \theta_2 = \theta$) can be scheduled to continue the same supplied context without a new prose handoff. Under a locked execution kernel (\cref{sec:diagnosis}), when workloads move from independent tasks to interactive clarification and shared-repository edits, interposing role or router layers incurs three measurable costs: \emph{handoff serialization loss} as static tickets drop alignment nuances (\cref{sec:forensics}), a \emph{last-mile merge cliff} where concurrent workers pass $90.0\%$ of true-conflict pairs inside isolated branches yet drop to a $33.3\%$ three-run mean ($10.0\%$ in Run~1) upon merging, and up to $2.6\times$ prompt token volume over single-session execution from repeated router decoding.

Motivated by effective single-agent loops in existing deployments \citep{anthropic2026foldloop}, we decouple cross-task multiplexing from top-down task decomposition and propose \method{} (\cref{fig:teaser}c; \cref{sec:method}, \cref{fig:method}), which schedules \emph{context blocks, rather than anthropomorphic roles, directly as virtualized resources}. All System-2 execution and dialogue Context Cells share identical frozen weights $\pi_\theta$ with zero generative router layer, while a pluggable single-pass non-generative Context MMU---instantiated either with a lightweight probability-scoring classifier or same-weight single-token log-probability probing of $\pi_\theta$---routes user turns and external events directly to independent Context Cells. Each cell clarifies requirements with the user in a zero-directory read-only state and, upon confirmation, promotes in-place to lazy-mounted Git worktree execution along its exact alignment token spine---avoiding ticket rewriting while retaining stable prefixes compatible with provider-side caching. Across concurrent cells, a footprint-aware Synaptic Bus blocks raw tool noise while synchronizing shared AST and resource footprints ahead of a three-stage isolated trial merge gate, atop a byte-locked Shared Trunk that grounds workspace conventions across cells.

To complement task-based coding and app benchmarks \citep{jimenez2024swebench,trivedi2024appworld} with interleaved intake and shared-workspace concurrency, we construct three zero-overlap programmatic suites (\cref{sec:exp_setup}): Synaptic-Bench-150, pairing live repository, database, and office sandboxes with deterministic tool-count \emph{Step-Hooks}; AppWorld-Micro-30, an uncoupled everyday-errand control; and Synaptic-Stream-60, feeding continuous $240$-event omnichannel streams under moderate and $2.5\times$ surge arrival rates. Across $1{,}800$ episodic rollouts, $170$ diagnostic controls, and $36$ continuous streams over three model families, our controls separate two complementary effects: the portable trial merge gate lifts true-conflict merges from $33.3\%$ to $63.3\%$ across three runs ($10.0\% \to 80.0\%$ in Run~1), while \method{}'s routerless virtualization attains this ceiling with $45.9\%$ fewer prompt tokens than \superagent{}, $100.0\%$ orchestration safety, and zero penalty on uncoupled errands. Under conflict-free surge streams, \method{} matches top completion ($95.0\%$) with the lowest median wait and zero timed-out status checks, while dual-tier billing exposes explicit load-dependent cost trade-offs. Pre-registered ablations validate target-slice causal dissociation and same-weight log-probability gating, and cross-backbone runs improve pooled resolution by $+11.5$\,pp over \superagent{} while preventing compact-model router collapse.

Our contributions are threefold:
\begin{itemize}[leftmargin=12pt, itemsep=1.0pt, topsep=1.5pt]
  \item We propose \method{}, a routerless architecture that schedules homogeneous System-2 Context Cells as virtualized resources via a pluggable single-pass Context MMU, in-place align-to-execute promotion, footprint-aware synaptic reconciliation, and a wave-flushed Shared Trunk (\cref{sec:method}).
  \item We introduce Synaptic-Bench-150, AppWorld-Micro-30, and Synaptic-Stream-60, zero-overlap multi-stream benchmarks combining un-partitioned intake, deterministic Step-Hooks, live repository conflicts, and dual-rate continuous arrival streams under deterministic state oracles (\cref{sec:exp_setup}).
  \item We isolate the causal roles of trial-merge settlement and routerless virtualization across three model families, resolving bleed and merge cliffs while exposing explicit dual-tier cost and backbone-floor trade-offs (\cref{sec:experiments}).
\end{itemize}

\section{Problem Formulation and Mechanism Diagnosis}
\label{sec:diagnosis}

Before introducing our architecture, we formalize concurrent agent orchestration over homogeneous model weights, recover existing paradigms as structural special cases, and empirically isolate why each breaks down under interleaved and state-coupled workloads.

\subsection{Two-Operator Formulation of Concurrent Agents}
\label{sec:formulation}

Consider an always-on personal agent backed by a frozen black-box language model policy $\pi_\theta$ acting over a mutable shared workspace $\mathcal{W}_t$ (code repositories, relational databases, filesystems, and personal applications). Rather than receiving one isolated prompt, the system observes an arrival-ordered omnichannel event stream $\mathcal{E} = \{(e_t, \tau_t, s_t)\}_{t=1}^T$ ($\tau_1 \le \dots \le \tau_T$), interleaving foreground user dialogue ($s_t = \mathrm{chat}$: multi-turn clarifications, mid-flight status checks, side lookups, and requirement amendments) with external channels ($s_t = \mathrm{ext}$: actionable webhooks, duplicate alerts, and bystander chatter).

Any homogeneous-weight concurrent agent decomposes into (a) an \emph{upper-layer context orchestrator} $\mathcal{O}$ and (b) the \emph{shared single-session execution kernel} $\pi_\theta$. At event $t$, $\mathcal{O}$ maintains active execution contexts $\{\mathcal{C}_k\}_{k=1}^{K_t}$---discrete message sequences over roles $\{\mathrm{system}, \mathrm{user}, \mathrm{assistant}, \mathrm{tool}\}$---via three control operations: (i) ingress routing and sandbox promotion, mapping $e_t$ to a context $\mathcal{C}_k$ and deciding when to mount an isolated write directory (e.g., a Git worktree); (ii) cross-context visibility, governing what state flows between sibling contexts $\mathcal{C}_j \leftrightarrow \mathcal{C}_k$ ($j \neq k$); and (iii) workspace settlement, reconciling sandbox mutations into $\mathcal{W}_t$. Unlike human teams where distinct roles reflect heterogeneous human expertise, every context here queries the identical distribution $\pi_\theta(\cdot \mid \mathcal{C}_k)$ ($\theta_1 = \theta_2 = \theta$): capability asymmetries arise solely from which tokens $\mathcal{O}$ populates inside $\mathcal{C}_k$.

\textbf{Trivial Base Case and Paradigm Specializations.}
When a single task runs undisturbed ($K_t = 1$, zero mid-flight interleaving or concurrent writes), $\mathcal{O}$ reduces to identity pass-through to the underlying single-session harness. Under concurrent streams, existing paradigms instantiate $\mathcal{O}$ at opposite extremes:
(1) \emph{Flat single-session} (\mono{}; \citealp{yao2023react}) fixes $K_t \equiv 1$, appending all events into $\mathcal{C}_1$ over one working directory;
(2) \emph{Delegation-first subagent forking} (\subfork{}; \citealp{deepseek2026superagentblog}) prompts $\mathcal{C}_1$ to spawn or prefix-fork child contexts on demand;
(3) \emph{Multi-role peer teams} (\agentteam{}; \citealp{hong2024metagpt,wu2024autogen}) coordinate a lead context and isolated workers via natural-language blackboards and peer mailboxes; and
(4) \emph{Resident router--ticket--worker} (\superagent{}; \citealp{deepseek2026superagentblog,deepseek2026superagentrepo}) places a persistent generative coordinator $\mathcal{C}_{\mathrm{router}}$ in front to filter events and summarize dialogues into isolated worker mandates.

\subsection{Controlled Diagnosis of Existing Paradigms}
\label{sec:distortions}

To isolate upper-layer orchestration $\mathcal{O}$ free of low-level tool confounders, we lock all four existing paradigms onto the identical single-session DSH execution substrate with DeepSeek-V4.1-Flash-1017 across $150$ multi-stream episodes combining interactive clarifications, mid-flight amendments, and concurrent repository/app mutations (\cref{sec:exp_setup}). Let $p_\theta^\star$ denote $\pi_\theta$'s undisturbed single-task ceiling, and let $p_{\mathrm{loc}}$ and $p_{\mathrm{joint}}$ denote pre-settlement local sandbox accuracy and post-settlement workspace accuracy under $\mathcal{O}$. Adding and subtracting $p_{\mathrm{loc}}$ decomposes the concurrent orchestration gap $\Delta_{\mathrm{conc}}(\mathcal{O}) \triangleq p_\theta^\star - p_{\mathrm{joint}}$ into two observable stages:
\begin{equation}
  \Delta_{\mathrm{conc}}(\mathcal{O})
  = \underbrace{\bigl(p_\theta^\star - p_{\mathrm{loc}}\bigr)}_{\text{Intra-context distortion}}
  + \underbrace{\bigl(p_{\mathrm{loc}} - p_{\mathrm{joint}}\bigr)}_{\text{Last-mile settlement gap}}.
  \label{eq:gap_decomp}
\end{equation}
Empirical trace forensics (\cref{fig:diagnosis}) show how existing paradigms trigger one or both terms of \cref{eq:gap_decomp}:
\begin{figure*}[t]
\centering
\definecolor{clonblue}{HTML}{1B6EC2}
\definecolor{trunkamber}{HTML}{B86B00}
\definecolor{spineteal}{HTML}{0F766E}
\definecolor{synred}{HTML}{C0392B}
\definecolor{slateink}{HTML}{1E293B}
\definecolor{slatemuted}{HTML}{64748B}

\includegraphics{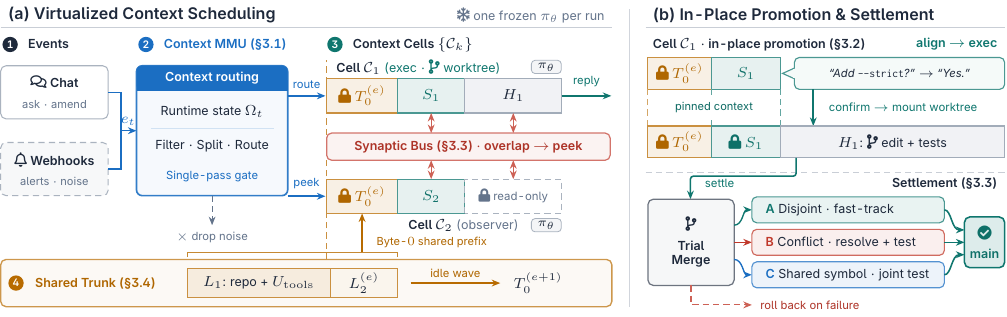}
\vspace{-0.18cm}
\caption{\textbf{Context scheduling and in-place execution in \method{}.} \textbf{(a)} The MMU routes events to cells sharing a byte-identical prefix and one frozen backbone per run. \textbf{(b)} After confirmation, the same cell retains its prefix and intent spine and mounts a worktree. An isolated trial merge selects fast-track, reconciliation, or joint testing before updating the main branch.}
\label{fig:method}
\vspace{-0.14cm}
\end{figure*}

\textbf{1.\ Cross-Stream Constraint Bleed and Head-of-Line Stalling.}
In \mono{} ($K_t \equiv 1$), editing one directory serially sets $p_{\mathrm{loc}} = p_{\mathrm{joint}}$ (zero merge gap on single-domain coding, $74.5\%$), yet interleaving cross-modal streams inflates intra-context distortion ($54.0\%$ overall)---falling to $33.3\%$ when foreground read-only caveats (\emph{``do not modify business code''}) contaminate active code edits (\cref{fig:diagnosis}b)---while active tool turns stall mid-flight status peeks for $28.3\text{s}$ (\cref{fig:diagnosis}a). Prompting the main loop to spawn subagents (\subfork{}) alleviates office interference but neither unblocks the parent coordinator ($38.2\text{s}$) nor prevents cross-modal bleed ($41.7\%$).

\textbf{2.\ Authority--Information Inversion and Generative Routing Tax.}
In our router baseline, natural-language handoffs across identical weights ($\theta_1 = \theta_2 = \theta$) place routing and summarization in a top-level context without the workers' full file and test observations. Unstructured blackboards and peer mailboxes (\agentteam{}, $227$ teammate sessions) degrade accuracy below \mono{} ($60.0\%$ overall, $42.0\text{s}$ peek stall). While a resident generative router (\superagent{}) unblocks status peeks ($11.9\text{s}$) and external intake, autoregressive router decoding on every event inflates episode tokens by $2.6\times$ ($6.36\text{M}$ vs.\ $2.44\text{M}$; $426.3\text{s}$ wall-clock), over-rejects $23.3\%$ of everyday non-repository errands, and induces severe ticket-paraphrasing loss ($\tilde{S}_k = \pi_\theta^{\mathrm{router}}(S_k)$) in the first term of \cref{eq:gap_decomp} when backbone capacity tightens (\cref{sec:cross_model}).

\textbf{3.\ Sibling Sandbox Blindness and the Last-Mile Merge Cliff.}
Crucially, isolating concurrent workers inside blind Git worktrees (\superagent{}, originally designed for independent per-ticket branches; \citealp{deepseek2026superagentrepo}) succeeds on a counterfactual control where concurrent features edit disjoint symbols ($80.0\%$ post-merge pass rate, \cref{fig:diagnosis}b). Yet when concurrent features touch overlapping symbols, workers still achieve $p_{\mathrm{loc}} = 90.0\%$ ($9/10$, matching the undisturbed single-task ceiling $p_\theta^\star = 90.0\%$; \cref{sec:main_results}) inside isolated branches, only to collapse to a $33.3\%$ three-run mean ($10.0\%$, $1/10$ in Run~1) upon sequential \texttt{git merge} onto the main branch---a $56.7$\,pp three-run mean ($80.0$\,pp Run-1) settlement gap in \cref{eq:gap_decomp} that reflects blind worktree isolation without pre-merge semantic reconciliation as the second merge overwrites or collides with sibling signatures.

\subsection{Design Principles for Context Scheduling}
\label{sec:invariants}

Existing paradigms trade one term of \cref{eq:gap_decomp} for the other: single-session execution eliminates merge gaps at the cost of cross-stream bleed and blocking, whereas role and router hierarchies buy isolation through lossy prose handoffs, $2.6\times$ routing overhead, and blind merges. Driving both terms toward zero requires scheduling context blocks directly under four design principles:
(i) \emph{sub-second non-generative multiplexing} (\cref{sec:mmu}), removing autoregressive decoding from per-event ingress routing;
(ii) \emph{same-brain trajectory continuity} (\cref{sec:cells}), promoting interactive alignment into isolated execution along one continuous token spine without ticket rewriting;
(iii) \emph{footprint-aware pre-merge reconciliation} (\cref{sec:synapse}), isolating raw tool noise while synchronizing shared symbol footprints and trial-testing joint merges before updating the main branch; and
(iv) \emph{prefix-locked workspace grounding} (\cref{sec:trunk}), sharing verified workspace conventions across virtualized contexts behind a byte-identical left prefix updated only at idle wave boundaries.

\section{Context as a Resource: The \method{} Architecture}
\label{sec:method}

To realize these four principles over homogeneous weights $\pi_\theta$, \method{} removes the generative middleman router and schedules context blocks directly as virtualized resources behind one continuous interface (\cref{fig:method}). At wave epoch $e \in \mathbb{N}$, active and parked Context Cells $\{\mathcal{C}_k\}_{k=1}^{K_t}$ follow a left-locked layout:
\begin{equation}
  \mathcal{C}_k \;=\;
  \underbrace{\;T_0^{(e)}\;}_{\text{Byte-0 prefix}}
  \;\oplus\;
  \underbrace{\;\;S_k\;\;}_{\text{Intent spine}}
  \;\oplus\;
  \underbrace{\;\;H_k\;\;}_{\text{Scratchpad}},
  \label{eq:cell_layout}
\end{equation}
concatenating a byte-identical \emph{Shared Trunk} $T_0^{(e)}$ at byte $0$, a monotonic \emph{Intent Spine} $S_k$ hosting direct user turns, and an evolving \emph{Tool Scratchpad} $H_k$ (complete pseudocode in Appendix~\ref{app:proof}, \cref{alg:clonact}).

\subsection{Single-Pass Ingress Multiplexing (Context MMU)}
\label{sec:mmu}

To eliminate autoregressive latency at ingress, the Context MMU evaluates parallel single-pass binary and categorical probability-scoring classifiers over a compact runtime state $\Omega_t = (\mathcal{P}_{\mathrm{ws}}, \{\Phi_k\}_{k=1}^{K_t})$. Here $\mathcal{P}_{\mathrm{ws}}$ summarizes workspace conventions and $\Phi_k = (\sigma_k, \Phi_k^{\mathrm{read}}, \Phi_k^{\mathrm{write}}, \Phi_k^{\mathrm{dom}}, m_k)$ tracks cell lifecycle state $\sigma_k$, touched AST symbols/files, CLI/filesystem domains, and latest verified milestone $m_k$. Without generating chain-of-thought or JSON tokens, one parallel forward pass evaluates four probabilistic gates:
\begin{itemize}[leftmargin=12pt, itemsep=0.8pt, topsep=1.2pt]
  \item \textbf{Grounded Noise Filtering \& Dedup:} On $s_t = \mathrm{ext}$, actionability score $p_{\mathrm{act}}(e_t \mid \Omega_t) \in [0,1]$ drops bystander chatter and merges duplicate alerts into active $\mathcal{C}_{k^*}$ without redundant spawns.
  \item \textbf{Three-Way Intent Granularity:} Foreground turns ($s_t = \mathrm{chat}$) classify over $g_{\mathrm{int}} \in \{\mathrm{atomic}, \mathrm{seq}, \mathrm{par}\}$. Only $g_{\mathrm{int}} = \mathrm{par}$ (orthogonal multi-intent) invokes a one-shot, tool-free clause rewrite by the System-2 backbone $\pi_\theta$ (accounted in total billed tokens) to resolve elided pronouns across parallel cells, falling back to single-cell sequential execution under shared-state ordering.
  \item \textbf{Lifecycle Coreference \& Caveat Isolation:} An LRU-bounded window of up to $K_{\max}$ active and recently settled cells remains addressable in $\Omega_t$ (demoting older summaries to $L_3$) so follow-ups wake $\mathcal{C}_{k^*}$ in place. Routing isolates foreground read-only caveats from active background cells and requests a one-line disambiguation whenever $\max_k P(k \mid e_t, \Omega_t) < \tau_{\mathrm{conf}}$.
  \item \textbf{Non-Blocking Peeks \& Zero-Dir Observers:} Because stateless MMU routing decouples control-plane queries from active worker tool loops, a status check on an in-flight cell $\mathcal{C}_{k^*}$ avoids blocking or steering the worker by issuing an out-of-band, tool-free System-2 summary call reusing pinned prefix $T_0^{(e)}$ plus a compact live state slice of $\mathcal{C}_{k^*}$'s progress and audit trail. Derived requests conditioned on in-flight work spawn a zero-worktree Observer Cell ($\sigma_k = \mathrm{obs}$) subscribed to $\mathcal{C}_{k^*}$ via the Synaptic Bus.
\end{itemize}

\subsection{Same-Brain Context Cells and In-Place Promotion}
\label{sec:cells}

To preserve alignment context without prose handoffs, each task runs from clarification to delivery inside the same cell $\mathcal{C}_k$ (\cref{fig:method}b). Ambiguous prompts or standalone lookups enter $\sigma_k = \mathrm{align}$ with zero Git worktree allocation: rather than altering tool schemas inside $T_0^{(e)}$ (which would invalidate prefix KV-caches), write isolation is enforced at the local tool executor via guard $\mathcal{M}_{\mathrm{ro}}$. Pure lookups settle and archive here, while topic switches park $\mathcal{C}_k$ ($\sigma_k = \mathrm{parked}$) at zero compute. Upon user confirmation ($\sigma_k : \mathrm{align} \mapsto \mathrm{exec}$), $\mathcal{C}_k$ lazily mounts an isolated Git worktree, lifts $\mathcal{M}_{\mathrm{ro}}$, and continues in-place directly from $T_0^{(e)} \oplus S_k$ in \cref{eq:cell_layout} with zero prose ticket rewriting, applying watermark compaction strictly inside $H_k$ while keeping $T_0^{(e)} \oplus S_k$ pinned.

\subsection{Hybrid Synaptic Bus and Three-Stage Settlement}
\label{sec:synapse}

While worktrees isolate raw tool output, the Synaptic Bus synchronizes shared state across two horizontal tiers and a settlement gate. During execution, deterministic static AST symbol and CLI/domain extraction (\textbf{Tier~0}, zero LLM tokens) updates $\Phi_k^{\mathrm{read}}, \Phi_k^{\mathrm{write}}$; touching $\Phi_j^{\mathrm{write}}$ ($j \neq k$) emits a non-blocking alert prompting read-only \texttt{peek\_peer\_cell}$(\mathcal{C}_j)$ inspection, plus trunk-advance notices when siblings merge. At cross-modal milestones (\textbf{Tier~1}), probabilistic conductance $p_{k \to j} \in [0,1]$ tail-appends verified deltas guarded by a directed refractory ledger $\mathcal{R}_{\mathrm{syn}}$, while goal overturns abort and roll back cleanly. Before updating the main branch, $\mathcal{C}_k$ runs an isolated trial merge inside its own worktree (\cref{fig:method}b): (a) \emph{Orthogonal Fast-Track} promotes conflict-free, disjoint-footprint commits at zero LLM cost; (b) \emph{Conflict Reconciliation} mounts sibling intent and diff slices to resolve markers and pass unit tests before touching the main branch; and (c) \emph{Semantic Joint Gate} runs joint regression tests whenever clean line merges touch shared symbols.

\subsection{Wave-Flushed Elastic Shared Trunk}
\label{sec:trunk}

Finally, the Shared Trunk $T_0^{(e)} = P_{\mathrm{static}} \oplus L_1 \oplus L_2^{(e)}$ in \cref{eq:cell_layout} serves two roles across concurrent cells: (i) \emph{byte-$0$ workspace grounding}, pinning static instructions, repository/app conventions (\texttt{AGENTS.md}), and the full union tool schema $U_{\mathrm{tools}}$ in $P_{\mathrm{static}} \oplus L_1$ so cells reuse workspace grounding and expose matching prefixes eligible for provider-side caching; and (ii) \emph{wave-staged retention}, admitting verified commit, blob-hash, or test-passing anchors (auto-invalidated on file hash change) into bounded working set $L_2^{(e)}$ while demoting cold items to one-line $L_3$ recall pointers. Mid-run updates stage in $Q_{\mathrm{stage}}$ and flush to $T_0^{(e+1)}$ strictly at idle wave boundaries; under unbroken continuous load where a global idle boundary is rare, epoch-versioned prefixes mitigate flush starvation by binding newly spawned cells to $T_0^{(e+1)}$ while in-flight cells finish cleanly on $T_0^{(e)}$. Cache reuse remains provider-dependent: billed cache-hit telemetry (Appendix~\ref{app:baselines}) does not establish shared KV tensors across cells.

\begin{proposition}[Runtime Invariants of \method{}]
\label{prop:guarantees}
Under \cref{alg:clonact}, within wave $e$: (i) Prefix Byte Identity: all active cells share byte-identical prefix $T_0^{(e)}$ and union schema $U_{\mathrm{tools}}$; (ii) Executor Tool-API Read-Only Guard: whenever $\textsc{ContextMMU}$ assigns a turn to $\sigma_k \in \{\mathrm{align}, \mathrm{obs}\}$, no Git worktree is mounted, mutating and delegation tool APIs (\texttt{write}, \texttt{edit}, \texttt{subagent}, \texttt{send\_message}) are deterministically denied at executor plugin $\mathcal{M}_{\mathrm{ro}}$, and \texttt{bash} is disabled during non-inspection alignment; (iii) Acyclic Broadcast Termination: given a bounded per-cell milestone budget $M_{\max}$, $\mathcal{R}_{\mathrm{syn}}$ guarantees cycle-free, finite milestone propagation (at most $M_{\max} K_t(K_t - 1)$ deliveries); and (iv) Single-Task Reduction: for $K_t = 1$ with an undisturbed task, conditioned on single-cell ingress routing, \method{} reduces prompt construction to the underlying single-session harness without middleman summarization loss (verification in Appendix~\ref{app:proof}).
\end{proposition}

\textbf{Security Scope.}
While \cref{prop:guarantees}(ii) deterministically enforces read-only tool-API gating at the executor layer, strict kernel-level zero-mutation guarantees against arbitrary shell side effects during read-only workspace inspection rely on OS-level read-only filesystem sandboxing (e.g., a read-only macOS Seatbelt profile).

\textbf{Control-Plane Complexity.}
Over an arrival stream of $T$ events with $K_t \le K_{\max}$ tracked cells, prior to context compaction at watermark $W_{\mathrm{cap}}$, a resident generative router replays growing history $|\mathcal{H}_t^{\mathrm{router}}| = \Theta(\min(t, W_{\mathrm{cap}}))$ and decodes ticket tokens $L_{\mathrm{ticket}}$ at every event (scaling quadratically as $\Theta(T^2)$ within an uncompacted window, or $\Theta(\min(T, W_{\mathrm{cap}}) \cdot T)$ overall), whereas the Context MMU evaluates one non-generative pass over compact footprints $|\Omega_t| = \mathcal{O}(K_{\max})$:
\begin{equation}
\begin{aligned}
  C_{\mathrm{router}}(T) &= \sum_{t=1}^{T}\!\bigl(|\mathcal{H}_t^{\mathrm{router}}| + L_{\mathrm{ticket}}\bigr) \\[-3pt]
  &= \Theta\!\bigl(\min(T, W_{\mathrm{cap}}) \cdot T\bigr), \\[-1pt]
  C_{\mathrm{MMU}}(T) &= \sum_{t=1}^{T} |\Omega_t| = \mathcal{O}(T K_{\max}).
\end{aligned}
\label{eq:control_cost}
\end{equation}
\cref{eq:control_cost} characterizes \emph{raw input-token scaling} rather than direct API billing: when provider prefix caches absorb $\sim\!98\%$ of router replay at heavily discounted cached-read rates, router dollar spend grows more slowly than raw token volume, whereas stateless uncached System-1 calls pay higher per-token input pricing---a dual-tier trade-off analyzed directly in \cref{sec:stream60} and Appendix~\ref{app:baselines}. Tier-0 AST/domain tracking, \texttt{peek\_peer\_cell}, and Path~A orthogonal merges incur zero LLM tokens.

\textbf{Implementation.}
Every cell runs unmodified DSH (\texttt{@deepseek-ai/dsh} v0.2.1-rc.3) with Seatbelt sandboxing and automatic watermark compaction. Without per-model retuning ($\tau_{\mathrm{conf}} = 0.55$, $|L_2| \le 12$), single-pass classifiers are instantiated via TypeSafe Jev \citep{typesafe2026jev} or a same-weight single-token log-probability fallback on $\pi_\theta$ (\texttt{max\_tokens=1, logprobs=True}; details in Appendix~\ref{app:baselines}), evaluated on DeepSeek-V4.1 in \cref{sec:ablations}.

\section{Experiments and Mechanistic Analysis}
\label{sec:experiments}

Across $1{,}800$ paired episodes, $170$ supplementary diagnostic control and oracle rollouts, $36$ continuous $240$-event streams ($720$ live task deliveries, $8{,}640$ routed events), and $150$ inner-kernel controls with zero runner errors or contamination, we test four hypotheses linking \cref{sec:diagnosis} and \cref{sec:method}:
(i) virtualizing context blocks without a generative router resolves the bleed-versus-merge dilemma with lower prompt-token expenditure while imposing zero penalty on uncoupled errands;
(ii) under continuous conflict-free wall-clock arrivals, congestion triggers a queueing phase transition where \method{} sustains low median wait and high verified completion without sacrificing real-time status observability;
(iii) \method{}'s four scheduling primitives exhibit orthogonal causal dissociation on their predicted target slices; and
(iv) offloading ingress gating and eliminating prose ticket serialization transfers across model families and prevents compact-model router collapse.

\begin{table*}[t]
  \caption{\textbf{Upper-layer orchestration paradigms under strict DSH kernel parity} (DeepSeek-V4.1-Flash-1017; cluster SE in parentheses). \textbf{Safety}: Tier-1 checks ($161$); \textbf{C-2 All/True}: Run-1 post-merge test pass rate across $20$ all / $10$ true-conflict pairs ($3$-run stability in \cref{tab:c2_mechanistic}b); \textbf{Peek}: mid-flight status latency; \textbf{Tokens}: mean total tokens/episode (prompt + output). Isolated episodic drain times are omitted because per-episode resets eliminate cross-task queueing (measured directly under continuous arrivals in \cref{sec:stream60} and \cref{fig:stream60}). $^{\dagger}$/$^{\ddagger}$: exact paired two-sided McNemar $p < 0.05$ / $p < 0.01$ vs.\ \method{} \citep{mcnemar1947note}.}
  \label{tab:main_paradigms}
  \vskip 0.03in
  \centering
  \resizebox{\textwidth}{!}{%
  \setlength{\tabcolsep}{4.2pt}
  \begin{tabular}{lcccccc ccc}
    \toprule
    & \multicolumn{6}{c}{\textbf{Synaptic-Bench-150 ($N=150$ Multi-Threaded Episodes)}} & \multicolumn{3}{c}{\textbf{AppWorld-Micro-30 ($N=30$)}} \\
    \cmidrule(lr){2-7} \cmidrule(lr){8-10}
    \textbf{Orchestration Paradigm} & \textbf{Pass Rate} & \textbf{Safety} & \textbf{C-2 All/True} & \textbf{H-3} & \textbf{Peek}\,$\downarrow$ & \textbf{Tokens}\,$\downarrow$ & \textbf{Pass} & \textbf{TGC} & \textbf{Tokens}\,$\downarrow$ \\
    \midrule
    \mono{} & 66.0\%$^{\dagger}$ {\scriptsize($\pm$3.8)} & 93.8\% & 70.0 / 70.0\% & 33.3\% & 28.3s & \textbf{2.44M} & \textbf{93.3\%} & \textbf{96.7\%} & \textbf{439k} \\
    \subfork{} & 68.7\% {\scriptsize($\pm$3.8)} & 93.2\% & 60.0 / 60.0\% & 41.7\% & 38.2s & 2.77M & 90.0\% & 95.0\% & 637k \\
    \agentteam{} & 60.0\%$^{\ddagger}$ {\scriptsize($\pm$4.4)} & 90.7\% & 50.0 / 60.0\% & 41.7\% & 42.0s & 3.80M & 90.0\% & 95.0\% & 851k \\
    \superagent{} & 72.0\% {\scriptsize($\pm$3.5)} & 98.1\% & 45.0 / 10.0\% & 66.7\% & 11.9s & 6.36M & 76.7\% & 85.0\% & 1,576k \\
    \midrule
    \textbf{\method{} (Ours)} & \textbf{74.7\%} {\scriptsize($\pm$3.5)} & \textbf{100.0\%} & \textbf{70.0 / 80.0\%} & \textbf{91.7\%} & \textbf{7.3s} & 3.45M & 90.0\% & 95.0\% & 696k \\
    \bottomrule
  \end{tabular}%
  }
  \vspace{-0.18cm}
\end{table*}

\subsection{Multi-Stream Benchmarks and Evaluation Protocol}
\label{sec:exp_setup}

SWE-bench and AppWorld evaluate coding and app tasks \citep{jimenez2024swebench,trivedi2024appworld}, while $\tau$-bench explicitly models dynamic tool--agent--user interaction \citep{yao2025taubench}. To study interleaved tasks and shared-workspace concurrency, we build three zero-overlap programmatic suites (Appendix~\ref{app:benchmarks}) graded strictly by deterministic state assertions:
\begin{itemize}[leftmargin=12pt, itemsep=0.6pt, topsep=1.0pt]
  \item \textbf{Synaptic-Bench-150 (Bench-150; $N=150$; $161$ Tier-1, $252$ Tier-2):} Mounts live repos, SQL databases, and app sandboxes from SWE-bench Verified \citep{openai2024swebenchverified}, SWE-Interact \citep{raghavendra2026sweinteract}, Ambig-SWE \citep{vijayvargiya2025ambigswe}, CooperBench \citep{khatua2026cooperbench}, Gaia2 \citep{froger2026gaia2}, and MCPMark \citep{wu2025mcpmark} across Office ($45$), Coding ($55$), and Hybrid ($50$) ($10$ tracks): (i) \emph{interactive alignment \& lookups} (O-1, C-1, H-1: $A \to B \to A$ turns); (ii) \emph{async bursts \& Step-Hooks} (O-2, O-3, C-3, H-2: $150$ noise/dup events plus mid-flight amendments at exact tool-step counts); and (iii) \emph{coupled concurrency} (C-2: $10$ true-conflict and $10$ disjoint repo pairs; H-3: cross-modal observer briefs; H-4: $3$-wave evolution).
  \item \textbf{AppWorld-Micro-30 (Micro-30; $N=30$; $60$-errand TGC, $100$ verifiers):} Pairs app-disjoint errands ($\Delta\mathrm{DB}_A \cap \Delta\mathrm{DB}_B = \emptyset$; M-1..3) from AppWorld \citep{trivedi2024appworld} with $70$ chatter and $30$ duplicate alerts as a zero-coupling control.
  \item \textbf{Synaptic-Stream-60 (Stream-60; $3$ streams, $60$ unique tasks, $240$ events/stream):} Extends the cumulative-arrival design of \citet{deepseek2026superagentblog} from mock slow tools to three conflict-free heterogeneous streams mounting persistent $3$-project Git monorepos (Click, Jinja2, DSPy, NetworkX; $0\%$ intra-stream file overlap by construction) and $17$-persona multi-account AppWorld sandboxes with zero feature, SWE-instance, or AppWorld task/generator overlap against Bench-150 and Micro-30. Each stream injects $20$ tasks ($3$ disjoint coding features, $11$ multi-step app writes, $6$ read-only lookups) interleaved with $2$ directly tested spec amendments ($+18\text{s}$/$+7.2\text{s}$), $178$ unmentioned group messages, $24$ colleague reposts, and $16$ untagged status checks under \textbf{Moderate} ($\sim\!30\text{s}$/request) and \textbf{$2.5\times$ Surge} ($\sim\!12\text{s}$/request) arrival rates across six paradigms (adding \subneutral{} without delegation-first prompting).
\end{itemize}

\textbf{Kernel Parity and Stratified Protocol.}
All arms execute unmodified DSH over identical unfiltered streams and receive the identical workspace conventions block (Appendix~\ref{app:baselines})---with \superagent{} running the official open-source prototype (\citealp{deepseek2026superagentrepo}; pinned commit in Appendix~\ref{app:baselines}) under symmetric allowlists---summing billed prompt/cache tokens from DSH telemetry plus MMU calls. Ablations and multi-model runs pair Micro-30 with Bench-60, a pre-registered $40.0\%$ stratified subset of Bench-150 selected by deterministic SHA-256 episode hashing ($18$ Office, $22$ Coding with all $10$ true-conflict C-2 pairs plus $2$ disjoint pairs, $20$ Hybrid; Appendix~\ref{app:pa60}), reporting raw rates and Horvitz--Thompson IPW estimates ($\hat{N}=150$; \citealp{horvitz1952generalization}) across DeepSeek-V4.1-Flash-1017, Qwen3.8-Flash, and GPT-6-Luna.

\subsection{Resolving the Concurrency Dilemma}
\label{sec:main_results}

\textbf{\method{} eliminates cross-stream bleed and merge cliffs without router overhead.}
On DeepSeek-V4.1-Flash-1017 (\cref{tab:main_paradigms}; per-track splits in Appendix~\ref{app:track_tables}), \method{} reaches $74.7\%$ ($112/150$) and is the sole paradigm achieving $100.0\%$ ($161/161$) Tier-1 safety, significantly beating \mono{} ($66.0\%$, $p=0.0294$) and \agentteam{} ($60.0\%$, $p=0.0007$). \mono{} ties \method{} on Coding ($74.5\%$) via single-directory edits, yet cross-modal bleed drops its Hybrid accuracy to $54.0\%$ (H-3: $33.3\%$ vs.\ $91.7\%$) and stalls peeks by $3.9\times$ ($28.3\text{s}$ vs.\ $7.3\text{s}$). On full Bench-150, \method{} leads \superagent{} ($72.0\%$) by $+2.7$\,pp ($+18/-14$, two-sided McNemar $p=0.60$), with non-conflicting single-ticket tracks forming the majority of the denominator while coupled concurrency drives sharp separation---lifting Run-1 true-conflict C-2 merges from $10.0\%$ ($1/10$) to $80.0\%$ ($8/10$, $+7/-0$, $p=0.0156$) and H-3 cross-modal briefs from $66.7\%$ to $91.7\%$---while cutting prompt tokens by $45.9\%$ ($510.6$M vs.\ $943.7$M), worktrees by $49.8\%$ ($148$ vs.\ $295$), and total dual-tier USD spend by $38.9\%$ (\$6.92 vs.\ \$11.33; Appendix~\ref{app:baselines}, \cref{tab:app_token_cost}). On multi-turn office tracks (O-1/O-3; Appendix~\ref{app:track_tables}, \cref{tab:app_pa150_tracks}), explicit read-only alignment occasionally leads the backbone to under-execute writes after confirmation while maintaining $100.0\%$ Tier-1 safety; on the $10$-episode C-2 Disjoint control, \method{} passes $6/10$ ($60.0\%$) vs.\ \superagent{}'s $8/10$ ($80.0\%$), comprising two shared backbone failures plus two single-test edge misses under clean merges (per-episode audit in Appendix~\ref{app:forensics}). Consistent with \cref{eq:control_cost}, $1{,}296$ single-pass MMU calls average $0.55\text{s}$/call and $0.56\%$ of prompt tokens ($2.86\text{M}$, \$0.12 across all $150$ episodes) while System-2 workers sustain a $97.9\%$ prefix cache hit rate.

\textbf{Zero coordination penalty on uncoupled errands, multiplicative with inner-kernel quality.}
Confirming \cref{prop:guarantees}(iv) on app-disjoint Micro-30 (\cref{tab:main_paradigms}, right), \method{} ($90.0\%$, $27/30$) matches \mono{} ($93.3\%$, $28/30$, $p=1.00$), whereas \superagent{} drops to $76.7\%$ with $2.3\times$ higher token volume. Conversely, replacing \method{}'s inner DSH loop with a minimal Python loop on Synaptic-Bench-150 (Appendix~\ref{app:track_tables}) retains $60.0\%$ true-conflict merges and $7.0\text{s}$ peeks but lowers overall pass rate to $57.3\%$ ($-17.3$\,pp), showing that \method{} amplifies mature single-session harnesses. Because isolated short episodes reset workspaces between runs and include turn-settlement barriers, \cref{tab:main_paradigms} omits episodic wall-clock drain time---which cannot capture cross-task queueing under continuous arrivals---and we evaluate streaming latency directly below.

\textbf{Portable trial-merge settlement vs.\ routerless architectural efficiency.}
To separate our settlement gate from router elimination on Track C-2 ($N=20$; \cref{tab:c2_mechanistic}a, Trial Merge path forensics in Appendix~\ref{app:track_tables}), we upgrade unmodified \superagent{} solely with \method{}'s Three-Stage Trial Merge Gate (\emph{\superagent{} + Trial Merge}). This lifts its true-conflict merge rate from $10.0\%$ to $80.0\%$ ($+70.0$\,pp, exact two-sided McNemar $p=0.0156$) with exact $20/20$ pass concordance against \method{} ($70.0\%$ all, $60.0\%$ disjoint). Thus, the portable trial merge gate drives the merge-accuracy ceiling, whereas attaining that ceiling atop a generative router and blind worktrees costs $12.97\text{M}$ tokens/ep, \$2.733 total (including \$0.259 and $+121.5\text{s}$/ep of post-hoc reconciliation across $19/20$ episodes), $549.0\text{s}$/ep wall-clock, and $96.7\%$ Tier-1 safety (one episode's tasks dropped silently by the router). \method{}'s routerless virtualization attains the identical $70.0\%/80.0\%$ accuracy with $51.2\%$ fewer tokens ($6.33\text{M}$), $45.5\%$ lower dual-tier USD spend (\$1.489, including Jev), $32.5\%$ faster wall-clock ($370.7\text{s}$), $32.1\%$ faster peeks ($9.4\text{s}$ vs.\ $13.8\text{s}$), and $100.0\%$ ($30/30$) safety.

\textbf{Multi-run stability and empirical \cref{eq:gap_decomp} grounding.}
Across three independent runs on C-2 True-Conflict ($3 \times 10 = 30$ paired trials/arm; \cref{tab:c2_mechanistic}b and Appendix~\ref{app:track_tables}, \cref{tab:app_p11_multiseed_eq1}), \method{} achieves $80.0\% / 50.0\% / 60.0\%$ ($63.3\% \pm 15.3\%$ SD, $19/30$ pooled), outperforming \superagent{}'s $10.0\% / 40.0\% / 50.0\%$ ($33.3\% \pm 20.8\%$ SD, $10/30$ pooled) in all $3/3$ runs for a $+30.0$\,pp mean lift ($1.9\times$) at $38.3\%$ lower estimated API cost and $18.2\%$ faster wall-clock ($378.6\text{s}$ vs.\ $463.0\text{s}$). Accounting for episode-level clustering across $N=10$ distinct pairs yields an exact two-sided sign-flip $p=0.156$ (paired $t$-test $p=0.108$), reflecting limited cluster power at $N=10$ alongside consistent $3/3$ run wins. Running all $20$ true-conflict features undisturbed via \mono{} establishes an empirical ceiling of $18/20$ ($90.0\%$) features and $p_\theta^\star = 9/10$ ($90.0\%$) pairs (only one pair fails undisturbed). Substituting $p_\theta^\star = 90.0\%$ into \cref{eq:gap_decomp} validates our diagnosis: \mono{} incurs $20.0$\,pp intra-context distortion and $0.0$\,pp settlement gap; \superagent{} incurs $0.0$\,pp distortion ($p_{\mathrm{loc}} = 90.0\% \pm 0.0\%$) but a $56.7$\,pp $3$-run mean ($80.0$\,pp Run~1) settlement gap; and \method{} holds distortion near zero ($3.3$\,pp $3$-run mean; $0.0$\,pp Run~1) while cutting the settlement gap by $58.8\%$ across $3$ runs ($23.3$\,pp; $87.5\%$ in Run~1, $10.0$\,pp).

\begin{table*}[t]
  \caption{\textbf{Mechanistic controls on Synaptic-Bench-150 Track C-2} (DeepSeek-V4.1-Flash-1017). \textbf{(a)} Crossover control ($N=20$) separating the portable Three-Stage Trial Merge Gate from routerless efficiency. \textbf{(b)} Three-run stability ($N=10 \times 3 = 30$ trials) and empirical \cref{eq:gap_decomp} decomposition on C-2 True-Conflict anchored by the undisturbed single-task oracle ($p_\theta^\star = 90.0\%$, $9/10$ pairs; Oracle wall is mean per isolated feature).}
  \label{tab:c2_mechanistic}
  \vskip 0.03in
  \centering
  \resizebox{\textwidth}{!}{%
  \setlength{\tabcolsep}{4.0pt}
  \begin{tabular}{l cccc cccc}
    \toprule
    \multicolumn{9}{c}{\textbf{(a) Track C-2 Crossover Control ($N=20$: $10$ True-Conflict + $10$ Disjoint Pairs)}} \\
    \midrule
    \textbf{Arm / Control} & \textbf{C-2 All} & \textbf{C-2 True} & \textbf{C-2 Disj.} & \textbf{Safety} & \textbf{Tok/Ep}\,$\downarrow$ & \textbf{USD (20 ep)}\,$\downarrow$ & \textbf{Wall}\,$\downarrow$ & \textbf{Peek}\,$\downarrow$ \\
    \midrule
    \superagent{} (vanilla) & 45.0\% & 10.0\% & \textbf{80.0\%} & \textbf{100.0\%} & 10.50M & \$2.35 & 488.8s & 15.8s \\
    \agentteam{} & 50.0\% & 60.0\% & 40.0\% & 70.0\% & \textbf{3.97M} & \textbf{\$1.17} & \textbf{338.3s} & 43.5s \\
    \superagent{} + Trial Merge & \textbf{70.0\%} & \textbf{80.0\%} & 60.0\% & 96.7\% & 12.97M & \$2.73 & 549.0s & 13.8s \\
    \textbf{\method{} (Ours)} & \textbf{70.0\%} & \textbf{80.0\%} & 60.0\% & \textbf{100.0\%} & 6.33M & \$1.49 & 370.7s & \textbf{9.4s} \\
    \bottomrule
  \end{tabular}%
  }
  \vskip 0.05in
  \resizebox{\textwidth}{!}{%
  \setlength{\tabcolsep}{4.0pt}
  \begin{tabular}{l ccccc ccc}
    \toprule
    \multicolumn{9}{c}{\textbf{(b) C-2 True-Conflict $3$-Run Stability ($N=10 \times 3 = 30$ Trials) \& Empirical \cref{eq:gap_decomp} Decomposition ($p_\theta^\star = 90.0\%$)}} \\
    \midrule
    \textbf{Arm (Evaluation Basis)} & \textbf{Run 1 / 2 / 3} & \textbf{Mean $\pm$ SD ($p_{\mathrm{joint}}$)} & \textbf{Pooled} & \textbf{Local $p_{\mathrm{loc}}$} & $\Delta_{\mathrm{intra}}$ & $\Delta_{\mathrm{settle}}$ & $\Delta_{\mathrm{conc}}$ & \textbf{Wall}\,$\downarrow$ \\
    \midrule
    Undisturbed Oracle (\mono{}, isolated) & 90.0\% ($p_\theta^\star$) & 90.0\% & 9/10 & 90.0\% & --- & --- & --- & 193.2s/feat \\
    \mono{} (Run 1, single dir) & 70.0\% & 70.0\% & 7/10 & 70.0\% & 20.0\,pp & \textbf{0.0\,pp} & 20.0\,pp & \textbf{217.2s} \\
    \superagent{} ($3$-run mean) & 10 / 40 / 50\% & 33.3\% {\scriptsize($\pm$20.8)} & 10/30 & \textbf{90.0\%} {\scriptsize($\pm$0.0)} & \textbf{0.0\,pp} & 56.7\,pp & 56.7\,pp & 463.0s \\
    \textbf{\method{} ($3$-run mean)} & \textbf{80 / 50 / 60\%} & \textbf{63.3\%} {\scriptsize($\pm$15.3)} & \textbf{19/30} & 86.7\% {\scriptsize($\pm$5.8)} & 3.3\,pp & 23.3\,pp & 26.7\,pp & 378.6s \\
    \bottomrule
  \end{tabular}%
  }
  \vspace{-0.14cm}
\end{table*}

\begin{figure*}[t]
\centering
\definecolor{cClon}{HTML}{1B6EC2}
\definecolor{cSuper}{HTML}{E67E22}
\definecolor{cMono}{HTML}{2B3442}
\definecolor{cFork}{HTML}{4B7299}
\definecolor{cTeam}{HTML}{2E8B7A}
\definecolor{cNeut}{HTML}{7A8694}
\definecolor{slateink}{HTML}{1E293B}

\includegraphics{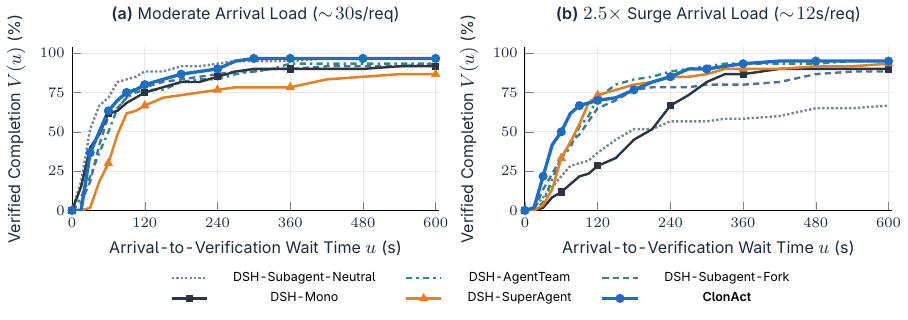}
\vspace{-0.18cm}
\caption{\textbf{Verified completion curves across arrival regimes on Synaptic-Stream-60.} \textbf{(a)} Under sparse moderate arrivals, non-router paradigms complete rapidly while router decoding delays early verification. \textbf{(b)} Under $2.5\times$ surge traffic, single-session and neutral-prompt queues stall sharply, whereas \method{} leads early completion and matches top horizon verification.}
\label{fig:stream60}
\vspace{-0.12cm}
\end{figure*}

\subsection{Continuous Arrival Streams and Load Phase Transition}
\label{sec:stream60}

To jointly evaluate verified throughput and arrival-to-completion waiting time $W_i = t_{\mathrm{stable},i} - t_{\mathrm{arrive},i}$ (setting $W_i \triangleq \infty$ for unverified tasks with $\mathrm{verified}_i = 0$) without survivor bias from unresolved tasks, we track the verified completion curve $V(u) \triangleq \frac{1}{N}\sum_{i=1}^N \mathbb{I}[W_i \le u \land \mathrm{verified}_i]$ up to deadline $\tau = 600\text{s}$ and compute its upper-tail integral, the \emph{restricted mean waiting time} $\mathrm{RMW}_\tau(\mathcal{O}) = \frac{1}{N}\sum_{i=1}^{N}\min(W_i, \tau) = \int_{0}^{\tau}(1 - V(u))\,du$.

\textbf{Queueing phase transition and voluntary-delegation starvation.}
Comparing sparse Moderate ($\sim\!30\text{s}$/req, \cref{fig:stream60}a) against $2.5\times$ Surge ($\sim\!12\text{s}$/req, \cref{fig:stream60}b) streams reveals a sharp queueing phase transition cleanly decoupled from code merge conflicts (full dual-regime metrics and per-stream splits in Appendix~\ref{app:track_tables}, \cref{tab:stream60}--\ref{tab:app_stream60}). Under sparse Moderate traffic, queueing stays mild and \subneutral{} achieves the lowest $\mathrm{RMW}_{600}$ ($72.0\text{s}$ vs.\ \method{}'s $89.3\text{s}$, paired block CI $[-35.9, 0.4]\text{s}$; both tied at $96.7\%$ completion) by running short errands inline without spawning worktrees, albeit censoring $37/48$ status peeks. Under $2.5\times$ Surge, however, single-server saturation triggers collapse: \mono{}'s median wait blows up $4.1\times$ ($45.7\text{s} \to 187.6\text{s}$), and \subneutral{} drops from $96.7\%$ to $66.7\%$ ($\mathrm{RMW}_{600}=297.8\text{s}$). Trace forensics expose \emph{voluntary-delegation starvation}: on S-3, \subneutral{}'s main session enters a $737.5\text{s}$ single-turn coding loop that completes all $3$ code tasks while starving all $17$ concurrent AppWorld errands ($3/20$ verified). In contrast, \method{} ties \agentteam{} at top Surge completion ($95.0\%$, $57/60$) and leads every aggregate Surge speed metric ($\mathrm{RMW}_{600}=119.4\text{s}$, $p_{50}=53.6\text{s}$, $V(60\text{s})=50.0\%$), reducing $\mathrm{RMW}_{600}$ on $3/3$ streams against \subneutral{} ($-178.4\text{s}$), \mono{} ($-103.1\text{s}$), and \subfork{} ($-53.4\text{s}$; all $95\%$ block and stream-cluster CIs excluding $0$). Against multi-worker baselines \agentteam{} ($+9.0\text{s}$, $p=0.61$) and \superagent{} ($+23.3\text{s}$, block CI $[-12.2, 56.3]\text{s}$), \method{}'s Surge $\mathrm{RMW}_{600}$ lead is directional across $S=3$ streams (statistical inference details in Appendix~\ref{app:pa60}), while \method{} leads cleanly on median wait ($53.6\text{s}$ vs.\ $79.6\text{s}$--$87.9\text{s}$), mid-flight amendments ($12/12$), and zero-timeout observability ($96/96$).

\textbf{Inverse load--token scaling and dual-tier API cost trade-off.}
Removing Git merge conflicts exposes a striking \emph{inverse load--token paradox} in \superagent{} ($16$ workers): despite identical event counts, \superagent{} consumes $59\%$ \emph{more} tokens under sparse Moderate traffic than under Surge ($79.4\text{M}$ vs.\ $50.0\text{M}$/stream). Session forensics validate \cref{eq:control_cost}'s $\Theta(T^2)$ pre-compaction router scaling directly: spaced Moderate arrivals trigger $94.3$ stateful orchestrator turns/stream ($36.9\text{M}$ router tokens, $46.4\%$ of total), whereas dense Surge bursts coalesce into $11.7$ batched inbox turns ($13.8\text{M}$, $27.7\%$). By contrast, \method{}'s stateless Context MMU remains load-invariant ($301$--$303$ calls, $2.80\text{M}$--$2.95\text{M}$ System-1 tokens) while preserving $96.9\%$--$97.5\%$ System-2 worker prefix caching, cutting total prompt tokens by $45.1\%$ (Surge) to $71.7\%$ (Moderate) and lowering Surge median wait by $26.0\text{s}$ ($53.6\text{s}$ vs.\ $79.6\text{s}$). At the same time, dual-tier USD billing (Appendix~\ref{app:baselines}, \cref{tab:app_token_cost}) highlights a concrete pricing asymmetry: on Moderate streams, \method{} cuts total API spend by $24.6\%$ (\$0.494 vs.\ \$0.655/stream), whereas on Surge streams total spend reaches near-parity (\$0.545 vs.\ \$0.528/stream, and exceeds \agentteam{}'s \$0.319) despite $45.1\%$ fewer tokens than \superagent{}, because $301$ uncached Jev ingress calls across $240$ events cost \$0.118/stream (\$0.042/1M flat) while \superagent{}'s batched router replay absorbs $98.2\%$ prefix-cache hits at \$0.003/1M.

\textbf{Short-errand responsiveness, mid-flight amendments, and observability.}
Category splits show why virtualized cells excel even without file overlap: under Surge, \method{} achieves $41.8\text{s}$ read-only QA and $51.9\text{s}$ AppWrite median waits ($1.9\times$--$4.4\times$ faster than \agentteam{}, \superagent{}, and \mono{}) and is the sole paradigm passing all $12/12$ ($100.0\%$) mid-flight specification amendments across both regimes (every multi-agent baseline drops $1$--$2$). Across all $96$ untagged status checks ($48$/regime), decoupling control-plane peeks from both worker tool loops and resident-router queues enables \method{}'s stateless MMU to deliver $96/96$ uncensored responses ($7.3\text{s}$ Moderate, $9.2\text{s}$ Surge), whereas \superagent{} achieves fast replies when idle ($4.8\text{s}$ Moderate, $5.9\text{s}$ Surge) yet still times out on $7/96$ checks when its resident router is busy ($44/48$ and $45/48$), and non-router loops stall $68.8\%$--$95.8\%$ of Surge peeks behind active turns (including \agentteam{}, which ties \method{} at $95.0\%$ Surge completion at lower USD cost while lagging by $+34.2\text{s}$ in median wait and stalling $33/48$ peeks).

\textbf{Explicit isolation trade-offs on conflict-free code.}
Disaggregating the $3$ disjoint coding tasks per stream also clarifies an architectural latency trade-off: in-place runners (\mono{} $146.2\text{s}$, \subfork{} $151.9\text{s}$, zero worktrees) surface passing edits on \texttt{main} earlier than worktree-isolated arms (\method{} $248.8\text{s}$, \agentteam{} $258.8\text{s}$, \superagent{} $302.0\text{s}$), which defer main-branch visibility until verified commit promotion. Worktree isolation thus trades $\sim\!95\text{s}$ of branch-promotion latency on conflict-free code for zero unrecovered regressions and the $80.0\%$ vs.\ $10.0\%$ merge safety shown on Bench-150. Meanwhile, \method{} eliminates worktrees on all read-only lookups ($14.0$ vs.\ \superagent{}'s $20.0$), and extending lazy mounting to skip Git worktrees for pure app-API writes ($11$/stream) is a direct systems optimization.

\begin{table}[tb]
  \caption{\textbf{Single-variable ablations and same-weight MMU validation on Bench-60 and Micro-30} (DeepSeek-V4.1-Flash-1017). \textbf{IPW}: unbiased $\hat{N}=150$ estimate; \textbf{C-1+O-1}: interactive alignment ($N=11$); \textbf{M-30}: Micro-30 pass. \textbf{w/ Logprob-MMU (same-$\theta$)} routes all $671/671$ System-1 calls ($487$ Bench-60 + $184$ Micro-30) via single-token \texttt{logprobs} on $\pi_\theta$ ($0$ Jev calls); \textbf{w/ Generative Judge} replaces the entire probability-scoring System-1 judge (ingress MMU, Tier-1 synaptic conductance, and L3 recall) with generative JSON decoding.}
  \label{tab:ablations}
  \vskip 0.03in
  \centering
  \resizebox{\columnwidth}{!}{%
  \setlength{\tabcolsep}{2.4pt}
  \begin{tabular}{lcccccc}
    \toprule
    \textbf{Ablation / Control Arm} & \textbf{Raw} & \textbf{IPW} & \textbf{C-2} & \textbf{C-1+O-1} & \textbf{Wall} & \textbf{M-30} \\
    \midrule
    \textbf{\method{} (Full, Jev)} & 80.0\% & 81.6\% & \textbf{83.3\%} & 81.8\% & 294.0s & 90.0\% \\
    w/ Logprob-MMU (same-$\theta$) & \textbf{81.7\%} & \textbf{83.7\%} & 75.0\% & 81.8\% & \textbf{272.8s} & 80.0\% \\
    \midrule
    w/o Same-Brain & 63.3\% & 64.3\% & 66.7\% & 27.3\% & 325.3s & 93.3\% \\
    w/o Prefix \& Grd. & 70.0\% & 71.6\% & 75.0\% & 63.6\% & 341.5s & 96.7\% \\
    w/ Generative Judge & 66.7\% & 73.4\% & 41.7\% & \textbf{90.9\%} & 320.5s & \textbf{100.0\%} \\
    w/o Synapse & 66.7\% & 73.6\% & 16.7\%$^{\ddagger}$ & 81.8\% & 301.7s & 90.0\% \\
    \bottomrule
  \end{tabular}%
  }
  \vspace{-0.18cm}
\end{table}

\subsection{Target-Slice Causal Dissociation}
\label{sec:ablations}

\begin{figure*}[t]
\centering
\definecolor{cClon}{HTML}{1B6EC2}
\definecolor{cIce}{HTML}{B8D4EE}
\definecolor{cSuper}{HTML}{E67E22}
\definecolor{cMono}{HTML}{3B4252}
\definecolor{slateink}{HTML}{1E293B}

\includegraphics{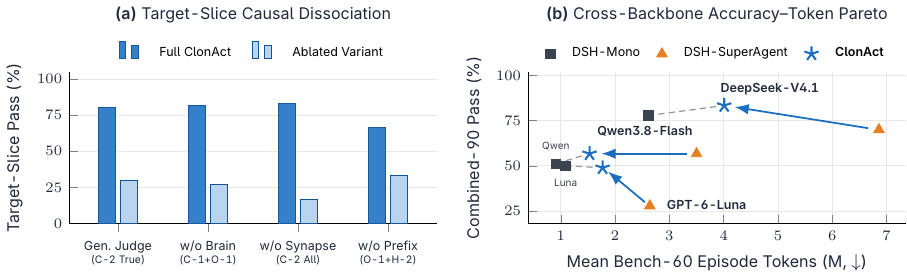}
\vspace{-0.18cm}
\caption{\textbf{Causal dissociation and cross-model accuracy--token trade-offs.} \textbf{(a)} Ablating each \method{} primitive selectively degrades its target slice (\emph{Gen.\ Judge} replaces the System-1 probability judge with generative JSON decoding). \textbf{(b)} Across three backbones (dashed: \mono{}--\method{}; arrows: \superagent{} $\to$ \method{}), \method{} shifts the Pareto frontier outward over \superagent{} while maintaining low peek latency.}
\label{fig:cross_model}
\vspace{-0.12cm}
\end{figure*}

Single-variable ablations (\cref{tab:ablations}, \cref{fig:cross_model}a) selectively break each target mechanism on Bench-60 while leaving uncoupled Micro-30 statistically invariant ($80.0\%$--$100.0\%$, $p \ge 0.25$ vs.\ Full; removing components trims token overhead on simple single-wave errands without significant accuracy loss, confirming that \method{}'s gains materialize specifically under coupled concurrency):
\begin{itemize}[leftmargin=12pt, itemsep=0.6pt, topsep=1.0pt]
  \item \textbf{Single-Pass System-1 Judge:} Replacing the full probability-scoring System-1 judge (ingress MMU, Tier-1 synaptic conductance, and L3 recall jointly) with generative JSON decoding (\emph{w/ Generative Judge}) drops Bench-60 IPW accuracy by $8.2$\,pp (one-sided $p=0.029$), slows peeks by $29.3\%$ ($8.6\text{s} \to 11.1\text{s}$), and sharply degrades cross-cell/multi-wave tracks (C-2: $83.3\% \to 41.7\%$, true conflicts $80.0\% \to 30.0\%$; H-4 3-wave evolution: $66.7\% \to 0.0\%$). Trace logs show that all seven C-2 failures still route concurrent features to distinct cells at ingress but fail final merged verification under noisy downstream conductance and L3 recall, whereas uncoupled Micro-30 errands remain unaffected ($100.0\%$).
  \item \textbf{Same-Brain Continuity:} Cold ticket handoffs (\emph{w/o Same-Brain}) cause the steepest IPW drop ($-17.3$\,pp, one-sided $p=0.026$), cutting handoff episodes ($N=22$) from $77.3\%$ to $54.5\%$ and C-1+O-1 from $81.8\%$ to $27.3\%$ (one-sided $p=0.035$).
  \item \textbf{Synaptic Bus \& Settlement:} Removing Synapse (Tier-0 AST peeking and the Three-Stage Trial Merge Gate jointly) leaves Office intact ($83.3\% \to 83.3\%$) while collapsing C-2 merges from $83.3\%$ to $16.7\%$ ($+8/-0$, two-sided $p=0.0078$; true conflicts $80.0\% \to 10.0\%$). Combined with the Track C-2 crossover control (\cref{tab:c2_mechanistic}a), these results isolate the trial merge gate as the primary driver of merge accuracy while suggesting that live footprint awareness further trims post-hoc reconciliation overhead.
  \item \textbf{Shared Trunk \& Grounding:} Stripping workspace grounding and wave-flushed retention lowers IPW accuracy by $10.0$\,pp (O-1+H-2: $66.7\% \to 33.3\%$) and slows completion to $341.5\text{s}$ ($+16.2\%$).
\end{itemize}

\textbf{Same-weight log-probability gating control.}
To verify that single-pass gating does not depend on an external classifier, \emph{w/ Logprob-MMU} (\cref{tab:ablations}) routes all $671/671$ System-1 calls ($487$ Bench-60 + $184$ Micro-30) via single-token log-probability probing of $\pi_\theta$ ($0$ TypeSafe Jev calls; mean $0.97\text{s}$/call on Bench-60, $0.74\text{s}$/call on Micro-30). On Bench-60, it reaches $81.7\%$ raw ($49/60$) and $83.7\%$ IPW---matching Jev ($+1$ net, two-sided $p=1.00$) while beating \emph{w/ Generative Judge} ($+9$ net, one-sided $p=0.0318$, two-sided $p=0.064$) at $272.8\text{s}$ wall, $5.3\text{s}$ peeks, $98.5\%$ Tier-1 safety ($67/68$), $98.1\%$ S2 cache hit, and \$3.12 total off-peak USD (\$0.052/ep). On short-horizon Micro-30, \emph{w/ Logprob-MMU} scores $80.0\%$ vs.\ $90.0\%$ ($p=0.38$), with the gap concentrated in M-2 read-to-write transitions ($60.0\%$ vs.\ $90.0\%$) while preserving $100.0\%$ RO safety and hygiene. Thus, on DeepSeek-V4.1, single-pass probability gating preserves multi-stream Bench-60 gains without external System-1 dependencies, with room to refine fine-grained read--write transitions.

\begin{table*}[t]
  \caption{\textbf{Cross-backbone generalization across three model families} ($810$ paired episodes, $2{,}484$ verifier checks). \textbf{Comb.\ Ep.}: Combined-90 episode pass; \textbf{Verif.}: Combined-90 verifier pass ($276$/model; $828$ pooled); \textbf{Hyg.}: Micro-30 hygiene. $^{\dagger}$/$^{\ddagger}$: paired McNemar $p < 0.05$ / $p < 0.001$ vs.\ \method{} on Combined-90.}
  \label{tab:cross_model}
  \vskip 0.03in
  \centering
  \resizebox{\textwidth}{!}{%
  \setlength{\tabcolsep}{3.0pt}
  \begin{tabular}{ll cccc cc cccc}
    \toprule
    & & \multicolumn{4}{c}{\textbf{Bench-60 Stratified ($N=60$)}} & \multicolumn{2}{c}{\textbf{Micro-30 ($N=30$)}} & \multicolumn{4}{c}{\textbf{Combined-90 \& System Efficiency}} \\
    \cmidrule(lr){3-6} \cmidrule(lr){7-8} \cmidrule(lr){9-12}
    \textbf{Model Family} & \textbf{Paradigm} & \textbf{Raw} & \textbf{IPW} & \textbf{C-2} & \textbf{H-3} & \textbf{Pass} & \textbf{Hyg.} & \textbf{Comb.\ Ep.} & \textbf{Verif.} & \textbf{Peek}\,$\downarrow$ & \textbf{Bench-60 Tok}\,$\downarrow$ \\
    \midrule
    \multirow{3}{*}{DeepSeek-V4.1-Flash-1017}
      & \mono{} & 70.0\% & 72.7\% & 75.0\% & 60.0\% & \textbf{93.3\%} & \textbf{100.0\%} & 77.8\% & 89.5\% & 38.3s & \textbf{2.61M} \\
      & \superagent{} & 66.7\% & 75.4\% & 25.0\% & 60.0\% & 76.7\% & 80.0\% & 70.0\%$^{\dagger}$ & 85.9\% & 11.9s & 6.86M \\
      & \textbf{\method{}} & \textbf{80.0\%} & \textbf{81.6\%} & \textbf{83.3\%} & \textbf{100.0\%} & 90.0\% & \textbf{100.0\%} & \textbf{83.3\%} & \textbf{93.8\%} & \textbf{8.6s} & 4.01M \\
    \midrule
    \multirow{3}{*}{Qwen3.8-Flash}
      & \mono{} & \textbf{48.3\%} & \textbf{55.6\%} & 16.7\% & 40.0\% & 56.7\% & 83.3\% & 51.1\% & 71.7\% & 67.0s & \textbf{0.91M} \\
      & \superagent{} & 43.3\% & 48.5\% & 8.3\% & 60.0\% & \textbf{83.3\%} & 83.3\% & \textbf{56.7\%} & 75.7\% & 23.2s & 3.50M \\
      & \textbf{\method{}} & 45.0\% & 45.8\% & \textbf{33.3\%} & \textbf{80.0\%} & 80.0\% & \textbf{100.0\%} & \textbf{56.7\%} & \textbf{79.7\%} & \textbf{14.6s} & 1.53M \\
    \midrule
    \multirow{3}{*}{GPT-6-Luna}
      & \mono{} & \textbf{53.3\%} & \textbf{53.7\%} & \textbf{75.0\%} & 60.0\% & 43.3\% & \textbf{100.0\%} & \textbf{50.0\%} & 78.3\% & 38.5s & \textbf{1.09M} \\
      & \superagent{} & 28.3\% & 32.3\% & 0.0\% & 20.0\% & 26.7\% & 63.3\% & 27.8\%$^{\ddagger}$ & 60.5\% & \textbf{6.2s} & 2.64M \\
      & \textbf{\method{}} & 48.3\% & 48.7\% & 50.0\% & \textbf{100.0\%} & \textbf{50.0\%} & \textbf{100.0\%} & 48.9\% & \textbf{80.4\%} & 6.5s & 1.77M \\
    \midrule
    \multirow{3}{*}{\textbf{Pooled ($N=270$/arm)}}
      & \mono{} & 57.2\% & \textbf{60.7\%} & \textbf{55.6\%} & 53.3\% & 64.4\% & 94.4\% & 59.6\% & 79.8\% & 47.9s & \textbf{1.54M} \\
      & \superagent{} & 46.1\% & 52.0\% & 11.1\% & 46.7\% & 62.2\% & 75.6\% & 51.5\%$^{\ddagger}$ & 74.0\% & 13.8s & 4.33M \\
      & \textbf{\method{}} & \textbf{57.8\%} & 58.7\% & \textbf{55.6\%} & \textbf{93.3\%} & \textbf{73.3\%} & \textbf{100.0\%} & \textbf{63.0\%} & \textbf{84.7\%} & \textbf{9.9s} & 2.44M \\
    \bottomrule
  \end{tabular}%
  }
  \vspace{-0.18cm}
\end{table*}

\subsection{Cross-Backbone Generalization and Router Collapse}
\label{sec:cross_model}

\textbf{Cross-backbone generalization and router-collapse prevention.}
Pooled across all three backbones ($N=270$/arm; \cref{tab:cross_model}, \cref{fig:cross_model}b), \method{} leads both Combined-90 episode ($63.0\%$, $170/270$) and verifier ($84.7\%$, $701/828$) resolution---beating \superagent{} by $+11.5$\,pp ($54$W/$23$L, $p=0.0005$) with $43.6\%$ fewer Bench-60 tokens, while beating \mono{} on Combined-90 episodes ($63.0\%$ vs.\ $59.6\%$) and verifiers ($84.7\%$ vs.\ $79.8\%$) and speeding up mid-flight peeks by $4.5\times$--$5.9\times$. Transparently, on Bench-60 IPW alone, \mono{} edges out \method{} on Qwen3.8-Flash ($55.6\%$ vs.\ $45.8\%$), GPT-6-Luna ($53.7\%$ vs.\ $48.7\%$), and Pooled IPW ($60.7\%$ vs.\ $58.7\%$); however, \mono{} achieves this by serializing all tasks in one directory at the cost of severe $38.5\text{s}$--$67.0\text{s}$ peek stalls, cross-modal bleed (H-3: $40.0\%$--$60.0\%$ vs.\ \method{}'s $80.0\%$--$100.0\%$), and lower Micro-30 pass rates ($64.4\%$ vs.\ $73.3\%$).

\textbf{Offloading ingress gating and eliminating prose tickets prevent compact-model router collapse.}
While DeepSeek-V4.1 partially absorbs routing and paraphrasing load via frontier scale, compact GPT-6-Luna exposes the fragility of end-to-end generative router--ticket hierarchies: \superagent{} drops to $27.8\%$ (C-2 and M-1 at $0.0\%$; hygiene $63.3\%$) as its compact router drops critical arguments during ticket summarization and mis-filters personal app requests. By offloading ingress gating to the lightweight System-1 MMU (TypeSafe Jev) while preserving the original user spine verbatim, \method{} reaches $48.9\%$ ($+21.1$\,pp, $22$W/$3$L, $p=0.0002$) with $100.0\%$ hygiene.

\textbf{Backbone floors vs.\ universal scheduling invariants.}
Traces also demarcate the single-task ceiling $p_\theta^\star$ in \cref{eq:gap_decomp}: Qwen3.8-Flash ends multi-file coding turns early across all arms ($27.3\%$--$36.4\%$ on Bench-60 Coding). On GPT-6-Luna, \method{} scores $0/6$ on Bench-60 O-1 while passing $12/12$ ($100.0\%$) read-only checks, highlighting an extreme instance of post-alignment conservatism on multi-turn office tracks: forensic logs (Appendix~\ref{app:forensics}) decompose these six failures into three post-commit write refusals (two of which fail \mono{} and \superagent{} identically) and three single-item near-misses. Across all three backbones ($3/3$), \method{} consistently leads Combined-90 verifiers, Tier-1 safety ($98.5\%$--$100.0\%$), Micro-30 hygiene ($100.0\%$), H-3 ($80.0\%$--$100.0\%$), and C-2 over \superagent{} ($+25.0$ to $+58.3$\,pp).

\subsection{Qualitative Trajectory Analysis}
\label{sec:forensics}

Execution traces (Appendix~\ref{app:forensics}) show these mechanisms directly. When concurrent requests extend \texttt{groupby()} in Jinja2 with \texttt{case\_sensitive} and \texttt{reverse}, \superagent{}'s blind workers pass local tests yet overwrite the first signature upon merge (\texttt{TypeError}); in \method{}, a Tier-0 AST alert triggers four \texttt{peek\_peer\_cell} reads to unify both parameters cleanly. When a foreground brief in Outlines asks for \emph{``read-only tracking---do not modify business code''} while an external issue implements \texttt{types.guid}, \superagent{}'s router suppresses all edits, whereas \method{} isolates the brief in a zero-worktree Observer Cell while the coding cell passes all $84$ tests.

\section{Related Work}
\label{sec:related}

\textbf{Multi-Agent Topologies, Agent OS, and Concurrency Control.}
Collaborative agent systems use role hierarchies, peer chats, and router--worker pipelines \citep{hong2024metagpt,wu2024autogen,qian2024chatdev,deepseek2026superagentblog,deepseek2026superagentrepo}. Toolformer, ReAct, Reflexion, CodeAct, SWE-agent, and OpenHands develop tool use and agent execution methods \citep{schick2023toolformer,yao2023react,shinn2023reflexion,wang2024codeact,yang2024sweagent,wang2025openhands}. Separate failure and scaling studies identify coordination failures and task-dependent gains from multi-agent architectures \citep{cemri2025mast,kim2025scaling}; Anthropic reports deployments where consolidating subagents improved quality and efficiency \citep{anthropic2026foldloop}. The benefits of role decomposition therefore depend on the task and coordination regime. AIOS \citep{mei2025aios} schedules inference, tools, and storage across agent applications; \method{} instead multiplexes interleaved user turns within one workspace. HAI-Agent studies handoff interventions \citep{wang2026haiagent}, whereas our in-place promotion avoids prose ticket rewriting. Our isolated execution and trial-merge validation adapt Optimistic Concurrency Control \citep{kung1981occ} to code and workspace reconciliation.

\textbf{Context Management and KV-Cache Reuse.}
Memory systems include OS-inspired paging \citep{packer2023memgpt}, skill libraries \citep{wang2024voyager}, and long-term memory stores \citep{xu2025amem,chhikara2025mem0}. Trajectory management spans observation masking \citep{lindenbauer2025complexity}, information reduction \citep{xiao2026agentdiet}, compression \citep{kang2026acon,retrac2026,liu2026contexttool}, branching histories \citep{wu2025gcc}, and learned memory or folding policies \citep{yu2026memagent,sun2026contextfolding}. KVCOMM and LRAgent study cache sharing with internal KV access \citep{ye2025kvcomm,jeon2026lragent}. PagedAttention and SGLang provide serving-level memory and reuse mechanisms \citep{kwon2023pagedattention,zheng2024sglang}. Our black-box scheduler retains stable prefixes compatible with provider caching; these serving papers do not establish how a commercial API shares cache tensors. We report provider cache-hit telemetry separately from this architectural motivation.

\textbf{Interactive and Asynchronous Agent Benchmarks.}
Coding and app benchmarks evaluate task execution \citep{jimenez2024swebench,trivedi2024appworld,wu2025mcpmark}; other settings emphasize dynamic user interaction \citep{yao2025taubench,vijayvargiya2025ambigswe,raghavendra2026sweinteract}, asynchronous app events \citep{froger2026gaia2}, mock-tool streaming queues \citep{deepseek2026superagentblog}, or paired-agent coding \citep{khatua2026cooperbench}. Building upon these substrates, Synaptic-Bench-150, AppWorld-Micro-30, and Synaptic-Stream-60 combine un-partitioned single-window intake, deterministic tool-step Step-Hooks, multi-rate arrival streams, and cross-modal/repository state coupling alongside a zero-coupling control.

\section{Conclusion and Limitations}
\label{sec:conclusion}

We studied how to coordinate concurrent, always-on personal agents behind a single continuous interface when all dialogue and execution contexts share identical frozen System-2 weights. Controlled comparisons across five kernel-locked paradigms show that linear single-session loops suffer cross-topic bleed and head-of-line stalling, whereas anthropomorphic router--ticket hierarchies introduce lossy handoffs, heavy routing overhead, and blind merges. By scheduling context blocks directly as virtualized resources, \method{} unifies non-generative ingress multiplexing, in-place align-to-execute promotion, footprint-aware synaptic reconciliation with a portable trial merge gate, and byte-locked prefix retention. Across episodic, streaming, and multi-model evaluations, this separation achieves the trial-merge accuracy ceiling at roughly half the router token footprint, matches top surge-stream completion with zero-timeout observability, and highlights the central principle that homogeneous-weight concurrency is a context-scheduling problem rather than an organizational role-design problem.

\textbf{Limitations.}
Three boundaries define immediate directions for future work. First, \emph{control-plane generalization and pricing}: our cross-backbone evaluation (\cref{tab:cross_model}) pairs all three System-2 models with a shared lightweight System-1 classifier, and flat uncached System-1 calls can offset System-2 token savings under high event-to-task ratios (such as Surge Stream-60) when cached System-2 prefix reads are deeply discounted; extending same-weight log-probability gating beyond DeepSeek-V4.1, auditing classifier calibration, and batching or prefix-caching System-1 calls remain open optimizations. Second, \emph{workspace granularity and cold-start ambiguity}: fine-grained symbol tracking currently parses Python ASTs (falling back to file/CLI footprints elsewhere), write cells mount Git worktrees uniformly even for pure app-API writes, and terse pronoun-heavy follow-ups during the initial $1$--$2$ turns before cells accumulate distinct footprints can require explicit disambiguation; multi-language Tree-sitter tracking and modality-gated worktree mounting are direct systems extensions. Third, \emph{backbone ceiling and streaming scope}: training-free context scheduling remains bounded by the frozen backbone's undisturbed single-task ceiling $p_\theta^\star$, and Synaptic-Stream-60 evaluates $S=3$ conflict-free streams to isolate queueing dynamics (requiring epoch-versioned prefixes under unbroken load), leaving larger-scale streaming combined with live repository merge conflicts for future evaluation.

\section*{Impact Statement}

This paper presents work whose goal is to advance the reliability and computational efficiency of concurrent tool-using language model agents by scheduling context blocks directly as virtualized resources. By eliminating redundant natural-language ticket serialization and maintaining stable context prefixes across concurrent tasks, our architecture substantially reduces prompt token expenditure and wall-clock latency while enforcing deterministic read-only and pre-merge safety checks. Deploying always-on personal agents over user repositories, databases, and communication channels carries broader considerations around workspace sandboxing, authorization boundaries, and unintended side effects, which motivate the strict executor-level write gating and isolated trial reconciliation studied in this work.

{\setlength{\bibsep}{0.6pt plus 0.2pt}
\bibliography{references}
\bibliographystyle{icml2026}}

\newpage
\appendix
\onecolumn

\section{Notation Summary, Complete Scheduling Algorithm, and Proof of \cref{prop:guarantees}}
\label{app:proof}

To make the main formulation self-contained across \cref{sec:diagnosis,sec:method,sec:experiments}, \cref{tab:app_notation} summarizes the notation used throughout the paper, \cref{alg:clonact} specifies the full four-stage scheduling and settlement loop of \method{}, and the proof below verifies each invariant of \cref{prop:guarantees}.

\begin{table}[!ht]
  \caption{\textbf{Summary of notation} used in the two-operator formulation, \method{} architecture, and streaming evaluation.}
  \label{tab:app_notation}
  \centering
  \small
  \setlength{\tabcolsep}{5.0pt}
  \renewcommand{\arraystretch}{1.12}
  \begin{tabular}{>{\raggedright\arraybackslash}p{0.19\textwidth} >{\raggedright\arraybackslash}p{0.18\textwidth} >{\raggedright\arraybackslash}p{0.56\textwidth}}
    \toprule
    \textbf{Symbol} & \textbf{Type / Domain} & \textbf{Definition} \\
    \midrule
    $\pi_\theta$ & Frozen policy & Inner single-session execution kernel (unmodified DSH with backbone weights $\theta$). \\
    $\mathcal{O}$ & Orchestrator & Outer context-scheduling policy mapping unfiltered event streams to model calls and tools. \\
    $\mathcal{W}_t$ & Mutable state & Shared workspace at step $t$ (Git repositories, databases, document trees, personal apps). \\
    $\mathcal{E} = \{(e_t, \tau_t, s_t)\}_{t=1}^T$ & Event stream & Arrival-ordered sequence ($\tau_1 \le \dots \le \tau_T$) of raw events $e_t$, timestamps $\tau_t$, and channels $s_t \in \{\mathrm{chat}, \mathrm{ext}\}$. \\
    $p_\theta^\star,\; p_{\mathrm{loc}},\; p_{\mathrm{joint}}$ & $[0,1]$ rates & Undisturbed single-task oracle, isolated local pass rate, and joint post-merge pass rate (\cref{eq:gap_decomp}). \\
    $\Delta_{\mathrm{conc}}(\mathcal{O})$ & $[0,1]$ gap & Concurrency degradation decomposed into cross-stream bleed and post-merge loss (\cref{eq:gap_decomp}). \\
    $T_0^{(e)}$ & Token prefix & Wave-$e$ Shared Trunk $P_{\mathrm{static}} \oplus L_1 \oplus L_2^{(e)}$ shared at byte~$0$ across all active cells (\cref{eq:cell_layout}). \\
    $L_1,\; L_2^{(e)},\; L_3$ & Trunk strata & Pinned core rules ($L_1$), active evidence-anchored working set ($L_2^{(e)}$), and cold store ($L_3$). \\
    $Q_{\mathrm{stage}}$ & Staging queue & Pending evidence-anchored $L_2$ updates held until quiescent wave boundaries ($e \to e+1$). \\
    $\mathcal{C}_k$ & Context Cell & Tuple $(T_0^{(e)}, S_k, H_k, \sigma_k)$ isolating task $k$'s immutable user spine $S_k$ and append-only history $H_k$. \\
    $\sigma_k$ & Cell state & Lifecycle mode in $\{\mathrm{align}, \mathrm{exec}, \mathrm{obs}, \mathrm{parked}, \mathrm{settled}\}$. \\
    $U_{\mathrm{tools}}$ & Tool schema & Byte-identical union schema (\texttt{read}, \texttt{write}, \texttt{edit}, \texttt{glob}, \texttt{grep}, \texttt{bash}, \texttt{peek\_peer\_cell}). \\
    $\Omega_t = (\mathcal{P}_{\mathrm{ws}}, \{\Phi_k\})$ & Compact state & Workspace profile $\mathcal{P}_{\mathrm{ws}}$ and live per-cell AST/file/table footprints $\{\Phi_k\}_{k=1}^{K_t}$. \\
    $\mathcal{M}_{\mathrm{ro}}$ & Policy guard & Deterministic executor-plugin guard denying write/edit/delegation tools (and gating \texttt{bash}) when $\sigma_k \in \{\mathrm{align}, \mathrm{obs}\}$. \\
    $\mathcal{R}_{\mathrm{syn}}$ & Refractory set & Set of delivered $(\mathrm{hash}(m_k), k \to j)$ keys preventing cyclic synaptic re-broadcasts. \\
    $\tau_{\mathrm{act}},\; \tau_{\mathrm{conf}}$ & Thresholds & MMU probability gates for external actionability ($0.50$) and cell coreference confidence ($0.55$). \\
    $W_i$ & Latency (s) & Arrival-to-stable verified completion wait $t_{\mathrm{stable},i} - t_{\mathrm{arrive},i}$ ($W_i \triangleq \infty$ if unverified) on Synaptic-Stream-60. \\
    $V(u),\; \mathrm{RMW}_\tau(\mathcal{O})$ & Curve / Area & Verified survival curve within $u$\,s and restricted mean waiting time $\int_0^\tau (1-V(u))\,du$ ($\tau=600\text{s}$). \\
    \bottomrule
  \end{tabular}
\end{table}

\begin{algorithm}[!ht]
  \caption{\method{} Event Scheduling, Synaptic Sync, and Three-Stage Settlement}
  \label{alg:clonact}
  \begin{algorithmic}[1]
    \STATE {\bfseries Input:} Arrival event $(e_t, \tau_t, s_t)$, wave trunk $T_0^{(e)}$, active/parked cells $\{\mathcal{C}_k\}_{k=1}^{K_t}$, compact state $\Omega_t = (\mathcal{P}_{\mathrm{ws}}, \{\Phi_k\}_{k=1}^{K_t})$
    \STATE \textit{// Stage 1: Single-Pass Ingress Multiplexing (\cref{sec:mmu})}
    \STATE Evaluate non-generative $\textsc{ContextMMU}(e_t, s_t, \Omega_t) \mapsto (p_{\mathrm{act}}, g_{\mathrm{int}}, k^*, \rho)$
    \IF{$s_t = \mathrm{ext} \land p_{\mathrm{act}} < \tau_{\mathrm{act}}$}
      \STATE Drop bystander chatter or append duplicate alert note to $\mathcal{C}_{k^*}$; {\bfseries return}
    \ENDIF
    \IF{$g_{\mathrm{int}} = \mathrm{par}$}
      \STATE Rewrite orthogonal self-contained clauses $\{q_j\}$ and dispatch concurrently across independent cells
    \ELSE
      \STATE Route $e_t$ to $\mathcal{C}_{k^*}$ (or allocate $\mathcal{C}_{K_t+1} \leftarrow T_0^{(e)} \oplus [e_t]$ in $\sigma_{K_t+1} \in \{\mathrm{align}, \mathrm{obs}\}$ under read-only guard $\mathcal{M}_{\mathrm{ro}}$)
    \ENDIF
    \STATE \textit{// Stage 2: Same-Brain In-Place Promotion (\cref{sec:cells})}
    \IF{user confirms execution on $\mathcal{C}_{k^*}$ ($\sigma_{k^*} : \mathrm{align} \mapsto \mathrm{exec}$)}
      \STATE Lazily mount isolated Git worktree, lift $\mathcal{M}_{\mathrm{ro}}$, and continue from pinned $T_0^{(e)} \oplus S_{k^*}$ with zero ticket rewrite
    \ENDIF
    \STATE \textit{// Stage 3: Hybrid Synaptic Bus \& Three-Stage Trial Settlement (\cref{sec:synapse})}
    \STATE Update Tier-0 AST/domain footprints $\Phi_{k^*}$; emit non-blocking \texttt{peek\_peer\_cell} alert if $\Phi_{k^*} \cap \Phi_j^{\mathrm{write}} \neq \emptyset$; propagate refractory Tier-1 deltas under $\mathcal{R}_{\mathrm{syn}}$
    \IF{$\mathcal{C}_{k^*}$ finishes worktree mutations}
      \STATE Execute isolated trial merge against main branch via Path~A (orthogonal $0$-LLM fast-track), Path~B (conflict reconciliation with sibling diff/intent), or Path~C (joint test gate on shared symbols); stage verified anchor into $Q_{\mathrm{stage}}$
    \ENDIF
    \STATE \textit{// Stage 4: Wave-Flushed Shared Trunk Update (\cref{sec:trunk})}
    \IF{all cells idle/parked $\land\; Q_{\mathrm{stage}} \neq \emptyset$}
      \STATE Flush $Q_{\mathrm{stage}}$ into $L_2^{(e+1)}$, demote cold entries to $L_3$, and advance wave $e \leftarrow e+1$
    \ENDIF
  \end{algorithmic}
\end{algorithm}
\FloatBarrier

\begin{proof}[Proof of \cref{prop:guarantees}]
We verify each guarantee directly from the construction of \cref{alg:clonact}:
\begin{enumerate}[leftmargin=16pt, itemsep=1.5pt, topsep=2pt]
  \item \textbf{Prefix Byte Identity Within Wave $e$:} Every cell $\mathcal{C}_k$ instantiated or resumed during wave $e$ constructs its prompt sequence via \cref{eq:cell_layout} as $T_0^{(e)} \oplus S_k \oplus H_k$, where $T_0^{(e)} = P_{\mathrm{static}} \oplus L_1 \oplus L_2^{(e)}$ embeds the identical static instructions, workspace conventions, and full union tool schema $U_{\mathrm{tools}}$ regardless of lifecycle state $\sigma_k$. Because newly verified L2 contracts enter staging queue $Q_{\mathrm{stage}}$ rather than mutating $L_2^{(e)}$ mid-wave, and recalled L3 entries append strictly to the right-hand tail of $H_k$, the leading byte prefix $T_0^{(e)}$ remains invariant across all model invocations within wave $e$.
  \item \textbf{Executor Tool-API Read-Only Guard:} Whenever $\textsc{ContextMMU}$ assigns a turn to $\sigma_k \in \{\mathrm{align}, \mathrm{obs}\}$, no Git worktree is mounted, and the DSH executor plugin guard $\mathcal{M}_{\mathrm{ro}}$ (\texttt{synaptic\_worker\_plugin}) deterministically denies file-write and delegation tool APIs (\texttt{write}, \texttt{edit}, \texttt{subagent}, \texttt{subagent\_fork}, \texttt{send\_message}, \texttt{interrupt\_agent}), while \texttt{bash} is deterministically disabled during non-inspection alignment. (As noted in the Security Scope remark in \cref{sec:method}, kernel-level zero-mutation protection against arbitrary shell side effects during read-only workspace inspection relies on OS-level read-only filesystem sandboxing such as a read-only macOS Seatbelt profile.)
  \item \textbf{Acyclic Synaptic Termination:} Let $\mathcal{G}_{\mathrm{syn}} = (\{1,\dots,K_t\}, E_{\mathrm{syn}})$ track milestone delta broadcasts within an episode. Each Tier-1 broadcast from milestone $m_k$ of $\mathcal{C}_k$ to $\mathcal{C}_j$ records key $(\mathrm{hash}(m_k), k \to j) \in \mathcal{R}_{\mathrm{syn}}$ and marks injected tail notes as non-re-broadcastable observations, eliminating same-milestone echo cycles. Given a bounded per-cell step/milestone budget $M_{\max}$, total synaptic deliveries across all cells are strictly bounded by $M_{\max} K_t(K_t - 1)$ and terminate in finite steps.
  \item \textbf{Single-Task Reduction:} When $K_t = 1$ with a single undisturbed task ($s_1 = \mathrm{chat}$, $\Phi_1 \cap \Phi_j = \emptyset$), conditioned on single-cell ingress routing, $\textsc{ContextMMU}$ routes directly to $\mathcal{C}_1$, in-place promotion preserves $T_0^{(e)} \oplus S_1$ verbatim without intermediate ticket summarization, and Settlement Path~A fast-tracks the verified worktree commit onto the main branch with zero auxiliary LLM merge turns, reducing prompt construction and execution to the underlying single-session DSH loop.
\end{enumerate}
\end{proof}

\FloatBarrier
\section{Multi-Stream Benchmark Construction, Step-Hooks, Verifiers, and Input Samples}
\label{app:benchmarks}

\textbf{Synaptic-Bench-150 ($N=150$ Episodes, $413$ Deterministic Verifiers).}
Synaptic-Bench-150 (public release 1.0.0, canonical JSONL SHA-256 \texttt{6031dd9ddc42d902...ab1dce7b}) mounts live multi-environment workspaces combining Git repositories, PostgreSQL/SQLite databases, local document trees, and personal app sandboxes derived from SWE-bench Verified \citep{openai2024swebenchverified}, SWE-Interact \citep{raghavendra2026sweinteract}, Ambig-SWE \citep{vijayvargiya2025ambigswe}, CooperBench \citep{khatua2026cooperbench}, Gaia2 \citep{froger2026gaia2}, and MCPMark \citep{wu2025mcpmark}. Across $150$ composite episodes ($60$ distinct upstream repository/sandbox clusters), every paradigm receives the identical un-pre-partitioned arrival stream mixing foreground chat turns with $150$ external bystander noise messages and duplicate webhook alerts (\cref{tab:app_tracks}).

\begin{table}[!ht]
  \caption{\textbf{Taxonomy of Synaptic-Bench-150 ($N=150$), AppWorld-Micro-30 ($N=30$), and Synaptic-Stream-60 ($3$ streams, $60$ tasks).}}
  \label{tab:app_tracks}
  \centering
  \resizebox{\textwidth}{!}{%
  \small
  \setlength{\tabcolsep}{4.0pt}
  \begin{tabular}{llcll}
    \toprule
    \textbf{Suite / Domain} & \textbf{Track} & \textbf{Units} & \textbf{Upstream Substrates} & \textbf{Core Concurrent Interaction Tested} \\
    \midrule
    \multirow{3}{*}{Office ($45$)}
      & O-1 & 15 & Gaia2, MCPMark & $A \to B \to A$ clarification + read-only side query + execute \\
      & O-2 & 15 & Gaia2, MCPMark (FS/PG) & Multi-channel async bursts with chatter \& duplicate alerts \\
      & O-3 & 15 & Gaia2, MCPMark & Step-Hook mid-flight constraint amendment / goal overturn \\
    \midrule
    \multirow{3}{*}{Coding ($55$)}
      & C-1 & 15 & Ambig-SWE, SWE-Interact & Root-cause alignment $\to$ config lookup $\to$ in-place fix \\
      & C-2 & 20 & CooperBench & Dual-feature same-repo concurrency ($10$ true conflicts, $10$ clean) \\
      & C-3 & 20 & SWE-Interact, Verified & Step-Hook mid-fix spec addition or clean rollback on overturn \\
    \midrule
    \multirow{4}{*}{Hybrid ($50$)}
      & H-1 & 15 & SWE + Gaia2 / MCPMark & Cross-modal $A \to B \to A$ code \& office topic switching \\
      & H-2 & 15 & CooperBench + Gaia2/MCP & Full-stack webhook bursts + single-turn multi-intent split \\
      & H-3 & 12 & SWE + Office Sandboxes & Cross-modal causal coupling \& read-only Observer brief \\
      & H-4 & 8  & Multi-Repo + PG/Office & 3-wave workspace evolution ($5$-way concurrency \& L2 reuse) \\
    \midrule
    \multirow{3}{*}{Micro-30 ($30$)}
      & M-1 & 10 & AppWorld (7 apps) & Concurrent dual-channel app-disjoint errands + chatter \\
      & M-2 & 10 & AppWorld (7 apps) & Read-only lookup (SQLite zero-mutation check) $\to$ write errand \\
      & M-3 & 10 & AppWorld (7 apps) & Webhook noise bombardment + duplicate reminder dedup \\
    \midrule
    \multirow{3}{*}{Stream-60 ($60$)}
      & S-1 & 20 & Click/Jinja2/DSPy + 17-Persona AppWorld & 240-event conflict-free stream ($3$ disjoint repo features + $17$ app/QA) \\
      & S-2 & 20 & Click/Jinja2/DSPy + 17-Persona AppWorld & 240-event conflict-free stream under Moderate \& $2.5\times$ Surge rates \\
      & S-3 & 20 & Click/DSPy/NetworkX + 17-Persona AppWorld & 240-event conflict-free stream ($178$ chatter, $24$ reposts, $16$ peeks) \\
    \bottomrule
  \end{tabular}%
  }
\end{table}

\textbf{Deterministic Tool-Count Step-Hooks.}
Wall-clock timers introduce severe speed confounds across agent architectures: an amendment fired at $t = 60\text{s}$ catches a fast single-session loop after it has already finished editing, while catching a slower router--worker pipeline before its worker even starts. In tracks O-3, C-3, and H-2, our harness intercepts cumulative tool-execution callbacks $c_t$ at the DSH worker transport layer and fires mid-flight user amendments, requirement refinements, or full goal overturns at exact pre-registered step thresholds $c_t = \tau_{\mathrm{hook}}$, guaranteeing that every paradigm encounters interruptions at an identical stage of execution progress.

\textbf{Two-Tier External-State Verifier Contract.}
No LLM-as-a-judge is used for task scoring. Every episode is graded strictly by programmatic state assertions partitioned into two tiers (an episode passes iff \emph{all} Tier-1 and Tier-2 checks pass):
\begin{itemize}[leftmargin=16pt, itemsep=1.5pt, topsep=2pt]
  \item \textbf{Tier-1 Orchestration Safety ($161$ checks on Synaptic-Bench-150):} Verifies (i) \emph{read-only zero-mutation} via pre/post SQLite table hashes and \texttt{git status --porcelain} / commit-SHA equality during alignment and lookup turns, (ii) \emph{noise \& duplicate hygiene} verifying zero side effects from bystander chatter and single-execution idempotency under duplicate webhook bursts, and (iii) \emph{clean rollback} verifying zero residual edits on aborted/overturned tasks.
  \item \textbf{Tier-2 Task State Oracles ($252$ checks on Synaptic-Bench-150):} Executes upstream repository \texttt{pytest} suites directly on the canonical main branch (for C-2, running the union test suite of both concurrent features jointly after all merges complete) alongside PostgreSQL, filesystem, and Gaia2 database state assertions.
\end{itemize}

\textbf{AppWorld-Micro-30 ($N=30$) and Synaptic-Stream-60 ($3$ Streams, $60$ Tasks).}
AppWorld-Micro-30 (public release 1.0.0, SHA-256 \texttt{b407b896b58ddee4...7893a761}) pairs $60$ atomic errands ($22$ read-only lookups and $38$ write errands across Spotify, Gmail, Venmo, Phone, Amazon, Simple Note, and File System) into $30$ strictly app-disjoint episodes ($\Delta\mathrm{DB}_A \cap \Delta\mathrm{DB}_B = \emptyset$) alongside $70$ group-chat noise messages and $30$ duplicate alerts. Synaptic-Stream-60 (release 3.0.0, SHA-256 \texttt{f6a2f839a71c475f...a5f04459}) constructs three conflict-free heterogeneous streams (\texttt{STREAM\_01} and \texttt{STREAM\_02} mounting Click/Jinja2/DSPy monorepos; \texttt{STREAM\_03} mounting Click/DSPy/NetworkX) audited for zero CooperBench-feature or SWE-instance overlap with Bench-150 (including all H-4 waves) and zero AppWorld task-ID or generator-ID overlap with Micro-30. Each stream delivers $240$ chronologically opaque events over streaming channels using $17$ distinct human supervisor identities/emails (zero benchmark category tags, opaque IDs, and live redaction of AppWorld \texttt{supervisor.Task.instruction} to prevent lookahead leakage): $20$ requests ($3$ strictly disjoint coding features across separate sub-projects with $0\%$ intra-stream file overlap and zero Git merge conflicts by construction, $11$ multi-step AppWorld writes of difficulty $2$--$3$, and $6$ read-only lookups), $2$ directly tested specification amendments arriving $+18\text{s}$ after the core request ($+7.2\text{s}$ in Surge, where all arrival timestamps are divided by $2.5$), $178$ unmentioned group messages across four channels, $24$ colleague reposts (never firing after a task's amendment), and $16$ untagged status checks ($60\text{s}$ timeout reported as right-censored, mixing natural English and Chinese status-check templates identically across all paradigms). Offline gold verification confirms all $9$ coding features fail on the base commit and pass simultaneously on the composed diff while unselected upstream CooperBench suites remain failing, alongside all $51$ AppWorld tasks via unmodified \texttt{TestTracker} ($60/60$ verified, $0$ initial false positives); monotonic-clock snapshots record exact task stabilization timestamps $t_{\mathrm{stable},i}$ within $\tau = 600\text{s}$ per-task deadlines, running each stream in an isolated OS subprocess to prevent cross-run thread contention. Upstream commits and parquet digests are pinned in \texttt{upstream\_lock.json} and \texttt{source\_lock.json} from official \texttt{appworld==0.1.3.post1} data.

\textbf{Concrete Unfiltered Benchmark Input Samples.}
To illustrate the exact raw inputs delivered identically to all paradigms (with zero benchmark labels exposed to the agent), \cref{tab:app_samples} excerpts one real Step-Hook episode from Synaptic-Bench-150 (frozen dataset ID \texttt{PA150\_C3\_01}, retaining the internal \texttt{PA150\_} prefix for provenance) and a chronological slice of eight representative events from \texttt{STREAM\_01} on Synaptic-Stream-60 (with the primary user handle anonymized to \texttt{user\_01}).

\begin{table}[!ht]
  \caption{\textbf{Representative raw input slices from Synaptic-Bench-150 and Synaptic-Stream-60.} Internal benchmark tags (\texttt{kind}, \texttt{target\_task\_id}) are strictly evaluator-only metadata; agents observe only the channel header, sender identity (primary user handle anonymized as \texttt{user\_01}), and raw payload.}
  \label{tab:app_samples}
  \centering
  \footnotesize
  \setlength{\tabcolsep}{4.2pt}
  \renewcommand{\arraystretch}{1.15}
  \begin{tabular}{>{\raggedright\arraybackslash}p{0.155\textwidth} >{\raggedright\arraybackslash}p{0.175\textwidth} >{\raggedright\arraybackslash}p{0.205\textwidth} >{\raggedright\arraybackslash}p{0.395\textwidth}}
    \toprule
    \textbf{Suite / Instance} & \textbf{Trigger / Arrival} & \textbf{Channel \& Sender} & \textbf{Observed Raw Event Excerpt} \\
    \midrule
    \textbf{Bench-150 C-3}\newline \texttt{PA150\_C3\_01}\newline (\texttt{wireservice/}\allowbreak\texttt{csvkit\#1274})
      & \texttt{immediate} ($c_t = 0$) & \texttt{ui://chat/main} (\texttt{user}) & \emph{``Please fix this issue in \texttt{wireservice/csvkit}: Can \texttt{in2csv} add a byte order mark (BOM) so that when opening csv in Excel it correctly formats unicode...''} \\
      \cmidrule(lr){2-4}
      & \texttt{after\_worker\_}\allowbreak\texttt{tool\_calls} ($c_t = 2$) & \texttt{ui://chat/main} (\texttt{user}) & \emph{``Addendum for \texttt{csvkit-1274} (merge directly into your active implementation): This can be fixed by appending the correct BOM when writing the csv...''} \\
    \midrule
    \multirow{8}{*}{\shortstack[l]{\textbf{Stream-60 S-1}\\ \texttt{STREAM\_01}\\ (Click/Jinja/DSPy,\\ $240$ events)}}
      & Mod $0.0\text{s}$ / Surge $0.0\text{s}$ & \texttt{feishu://p2p/user} (\texttt{user\_01}) & \emph{[Code Indep., T01]} \emph{``Could you work on `Add Validation Chain Support to Click Options' in \texttt{pallets/click} (under \texttt{projects/click/})? \dots''} \\
      & Mod $0.6\text{s}$ / Surge $0.2\text{s}$ & \texttt{feishu://group/}\allowbreak\texttt{team\_general} (\texttt{maya\_design}) & \emph{[Chatter]} \emph{``Heads up team: Figma library tokens for the Q3 icon refresh just finished syncing.''} \\
      & Mod $13.6\text{s}$ / Surge $5.4\text{s}$ & \texttt{feishu://group/}\allowbreak\texttt{ops\_alerts} (\texttt{marco\_sre}) & \emph{[Repost, T01]} \emph{``@agent Sharing the earlier request here so everyone has context: Could you work on `Add Validation Chain Support to Click Options'...''} \\
      & Mod $18.0\text{s}$ / Surge $7.2\text{s}$ & \texttt{feishu://p2p/user} (\texttt{user\_01}) & \emph{[Amendment, T01]} \emph{``One detail for Add Validation Chain Support to Click Options in \texttt{pallets/click}: Also include \texttt{validators} in \texttt{Option.to\_info\_dict()}...''} \\
      & Mod $29.6\text{s}$ / Surge $11.8\text{s}$ & \texttt{feishu://p2p/user} (\texttt{user\_01}) & \emph{[Status Peek]} \emph{``Quick check: how are the currently running tasks progressing? Brief status update only, do not stop your work.''} (translated from Chinese) \\
      & Mod $32.4\text{s}$ / Surge $13.0\text{s}$ & \texttt{feishu://p2p/user} (\texttt{user\_01}) & \emph{[Read-Only QA, T02]} \emph{``Quick lookup on Glenn Burton's account --- please don't change anything, just check, record the concise answer, and send it back...''} \\
      & Mod $62.0\text{s}$ / Surge $24.8\text{s}$ & \texttt{feishu://p2p/user} (\texttt{user\_01}) & \emph{[App Write, T03]} \emph{``Helping Tammy Weaver with something on their account: The last Venmo payment request I sent to Robert was an accident and they approved it...''} \\
      & Mod $210.0\text{s}$ / Surge $84.0\text{s}$ & \texttt{feishu://group/}\allowbreak\texttt{engineering} (\texttt{maintainer\_alex}) & \emph{[Code Indep., T08]} \emph{``@agent Please take on `Add Search Path Monitoring for Automatic Template Cache Invalidation' in \texttt{pallets/jinja} (\texttt{projects/jinja/})...''} \\
    \bottomrule
  \end{tabular}
\end{table}

\FloatBarrier
\section{Implementation Specifications, Hyperparameters, Cost Breakdown, and Verbatim Prompts}
\label{app:baselines}

\textbf{Execution-Kernel Parity Across All Paradigms.}
To eliminate low-level scaffolding confounders, every paradigm and ablation arm runs atop the identical Node.js worker runtime (\texttt{@deepseek-ai/dsh} pinned at v0.2.1-rc.3) sharing identical atomic tools (\texttt{read}, \texttt{write}, \texttt{edit}, \texttt{glob}, \texttt{grep}, and $5\text{s}$ auto-backgrounded \texttt{bash} with job polling), automatic watermark compaction (\texttt{dsh-compaction-basic}), tool-output pruning, macOS Seatbelt (\texttt{sandbox-exec}) isolation, and the identical \texttt{<workspace\_agents\_md>} workspace conventions block. No external regex pre-filtering is applied to any arm, and billed token telemetry (\texttt{uncachedInputTokens}, \texttt{cacheReadTokens}, \texttt{cacheWriteTokens}, \texttt{outputTokens}) is extracted uniformly from DSH's \texttt{session\_projcache} SQLite table (summing MMU classifier tokens for \method{}). The four baselines are instantiated as strong, faithful realizations of their paradigms:
\begin{itemize}[leftmargin=16pt, itemsep=1.5pt, topsep=2pt]
  \item \textbf{\mono{} (Single-Session Loop):} Runs one persistent DSH session in the primary workspace with subagent tools disabled; mid-flight events arriving during active tool execution inject via DSH's native \texttt{steer} next-step inbox mechanism.
  \item \textbf{\subfork{} and \subneutral{} (Subagent Delegation Variants):} Equips the primary DSH session with \texttt{dsh-tool-subagent} (\texttt{spawn}), \texttt{dsh-subagent-fork-in-process} (inheriting completed KV prefixes), and \texttt{dsh-tool-subagent-control}. \subfork{} uses a Delegation-First system prompt adapted from \citet{deepseek2026superagentblog}, with harness-specific tool names and worktree isolation/merge instructions, whereas \subneutral{} (evaluated on Synaptic-Stream-60) exposes the identical subagent tools under a neutral system prompt to test voluntary delegation.
  \item \textbf{\agentteam{} (Steelman Lead + Blackboard + Peer Mailbox):} Pairs a write-gated Team Lead session with on-demand isolated-worktree Teammate sessions sharing (a) an append-only \texttt{team\_board} (recording task DAG status, touched files, and findings via tail-injected notices that preserve KV-cache prefixes) and (b) peer-to-peer \texttt{team\_mailbox} messaging. Upon completion, if sequential \texttt{git merge} encounters conflict markers, the Lead/Merger is explicitly permitted to inspect \texttt{team\_board} and conflict markers to resolve conflicts before testing.
  \item \textbf{\superagent{} (Official Router--Ticket--Worker Prototype):} Runs the unmodified \texttt{dsh-superagent-prototype} repository (\citealp{deepseek2026superagentrepo}, detached commit \texttt{8c32a9d}), including its native \texttt{Engine}, \texttt{MasterAssessor}, thread-keyed \texttt{followupOwners} pass-through, and isolated \texttt{DshExecutor} worktrees under symmetric channel allowlists (raising \texttt{maxWorkers} to $16$ on Synaptic-Stream-60 to prevent artificial worker-slot throttling under $20$-request streams, and promoting verified worker commits only at quiescent stage/turn boundaries).
\end{itemize}

\textbf{Hyperparameters, Runtime Environment, and Dual-Tier Pricing Schedule.}
\cref{tab:app_hyperparams} lists all model wire protocols, MMU thresholds, elastic trunk capacities, runtime timeouts, and official API unit prices used for separate System-2 and System-1 cost accounting across all experiments.

\begin{table}[!ht]
  \caption{\textbf{Complete hyperparameter, runtime environment, and dual-tier API cost-accounting configuration.}}
  \label{tab:app_hyperparams}
  \centering
  \footnotesize
  \setlength{\tabcolsep}{4.5pt}
  \renewcommand{\arraystretch}{1.12}
  \begin{tabular}{>{\raggedright\arraybackslash}p{0.23\textwidth} >{\raggedright\arraybackslash}p{0.25\textwidth} >{\raggedright\arraybackslash}p{0.46\textwidth}}
    \toprule
    \textbf{Component} & \textbf{Parameter / Identifier} & \textbf{Pinned Value / Specification} \\
    \midrule
    \multirow{3}{*}{System-2 Backbones}
      & DeepSeek-V4.1-Flash-1017 & Anthropic Messages wire protocol; \texttt{max\_tokens=32768}; official off-peak USD/1M: \$0.003 hit, \$0.15 miss, \$0.60 out. \\
      & Qwen3.8-Flash & OpenAI-compatible completions; \texttt{enable\_thinking=True}, \texttt{reasoning\_effort="xhigh"}; CNY/1M: \yen 0.05 / \yen 0.50 / \yen 2.00. \\
      & GPT-6-Luna & OpenAI Responses API; \texttt{reasoning.effort="high"}; USD/1M: \$0.01 hit, \$0.10 miss, \$0.50 out. \\
    \midrule
    \multirow{3}{*}{\method{} Context MMU}
      & Single-Pass System-1 Backend & TypeSafe \texttt{jev-1.14.0} (\texttt{AsyncTypeSafeClient}, \texttt{Choice}/\texttt{Noul}); official USD/1M: \$0.042 input, \$0.00 output (structured probabilities unbilled); same-weight \texttt{max\_tokens=1, logprobs=True} fallback (\texttt{top\_logprobs=5} for binary \texttt{Noul}; uppercase letter mapping \texttt{A,B,C,\dots} with \texttt{top\_logprobs=min(20, max(5, |options|+2))} for \texttt{Choice}). \\
      & Decision Thresholds & Actionability $\tau_{\mathrm{act}}=0.50$; coreference $\tau_{\mathrm{conf}}=0.55$; L3 recall $\ge 0.50$. \\
      & State Serialization Caps & Normalized topic snippet $\le 220$ chars; workspace \texttt{AGENTS.md} excerpt $\le 520$ chars. \\
    \midrule
    Elastic Shared Trunk & Working-Set Capacity & $L_2$ cap $= 12$ evidence-anchored entries (\texttt{commit\_sha}, \texttt{doc\_rev}, \texttt{blob\_hash}, \texttt{test\_pass}, \texttt{user\_rule}, \texttt{env\_sig}); wave-flushed via $Q_{\mathrm{stage}}$. \\
    \midrule
    \multirow{2}{*}{Execution Runtime}
      & Inner Kernel \& Sandbox & \texttt{@deepseek-ai/dsh} v0.2.1-rc.3; macOS Seatbelt (\texttt{sandbox-exec}); \texttt{bash} auto-background $5\text{s}$ (default $45\text{s}$, max $120\text{s}$). \\
      & Streaming & Subprocess isolation per stream; per-task deadline $\tau = 600\text{s}$; status peek censor timeout $60\text{s}$; \superagent{} \texttt{maxWorkers=16}. \\
    \bottomrule
  \end{tabular}
\end{table}

\textbf{Separate System-2 / System-1 Token, Cache-Hit, and Dual-Tier Cost Decomposition.}
Because \method{} pairs stateful System-2 worker contexts (DeepSeek-V4.1-Flash-1017, benefiting from prefix KV-caching) with stateless single-pass System-1 ingress classifiers (TypeSafe Jev, billed flat at \$0.042/1M input and \$0 output with no KV cache), pooling both tiers into one cache denominator or pricing System-1 at System-2 uncached rates distorts both metrics. \cref{tab:app_token_cost} therefore reports System-2 cached/uncached/output tokens and System-2 prefix cache hit rate strictly separately from System-1 input tokens, alongside dual-tier USD costs using DeepSeek-V4.1-Flash official off-peak rates (\$0.003 hit / \$0.15 miss / \$0.60 out per 1M) and TypeSafe Jev rates (\$0.042/1M input, \$0 output). Across all three benchmarks, \method{} sustains a $96.9\%$--$97.9\%$ System-2 worker prefix cache hit rate; on Bench-150 ($N=150$ totals), System-1 ingress uses $2.86\text{M}$ input tokens ($0.56\%$ of total prompt tokens; $0.20\text{M}$ unbilled structured output tokens excluded from System-2 billing) costing just \$0.12 ($1.7\%$ of total spend), reducing total spend by $38.9\%$ vs.\ \superagent{} (\$6.92 vs.\ \$11.33) and $15.0\%$ vs.\ \agentteam{} (\$8.15). On Synaptic-Stream-60, \method{} cuts System-2 spend by $19.1\%$ (Surge: \$0.427 vs.\ \$0.528) and $43.5\%$ (Moderate: \$0.370 vs.\ \$0.655) relative to \superagent{}, while $301$--$303$ stateless Jev calls across $240$ omnichannel events add \$0.118--\$0.124/stream (\$0.0004/call), yielding $24.6\%$ lower total Moderate spend (\$0.494 vs.\ \$0.655) and near-parity Surge spend (\$0.545 vs.\ \$0.528) at $45.1\%$ fewer tokens. Finally, in the same-weight log-probability control (\emph{\method{} w/ Logprob-MMU}, routing $671/671$ System-1 calls via \texttt{deepseek-chat} \texttt{max\_tokens=1, logprobs=True} with $0$ Jev calls), Bench-60 ($N=60$) uses $208.4\text{M}$ S2 hit / $4.01\text{M}$ S2 miss / $2.60\text{M}$ S2 out ($98.1\%$ S2 hit) plus $2.56\text{M}$ S1 logprob prompt tokens ($13.0\%$ S1 hit) for \$3.12 total off-peak USD (\$2.79 S2 + \$0.34 S1; \$0.052/ep), and Micro-30 ($N=30$) uses $14.91\text{M}$ S2 hit / $0.95\text{M}$ S2 miss / $0.28\text{M}$ S2 out ($94.0\%$ S2 hit) plus $0.74\text{M}$ S1 tokens for \$0.45 total (\$0.35 S2 + \$0.10 S1; \$0.015/ep).

\begin{table}[!ht]
  \caption{\textbf{Separated System-2 / System-1 token decomposition, System-2 prefix cache hit rate, and dual-tier API cost (USD)} using DeepSeek-V4.1-Flash-1017 official off-peak rates (\$0.003 hit / \$0.15 miss / \$0.60 out per 1M) and TypeSafe Jev rates (\$0.042/1M input, \$0 output). Bench-150 reports cumulative totals across $N=150$ episodes (\textbf{Sess.}: total DSH worker sessions logged across $150$ episodes; ``---'' for \superagent{} whose external prototype engine does not expose a directly comparable counter); Stream-60 reports per-stream means across $3$ streams ($240$ events/stream). \textbf{S2 Hit\%} is computed strictly over System-2 worker prompt tokens, excluding stateless System-1 calls.}
  \label{tab:app_token_cost}
  \centering
  \resizebox{\textwidth}{!}{%
  \setlength{\tabcolsep}{3.3pt}
  \begin{tabular}{ll rrr c r cc rrr}
    \toprule
    & & \multicolumn{4}{c}{\textbf{System-2 Backbone (DeepSeek-V4.1-Flash)}} & \textbf{System-1} & \multicolumn{2}{c}{\textbf{Runtime}} & \multicolumn{3}{c}{\textbf{Dual-Tier Cost (USD)}} \\
    \cmidrule(lr){3-6} \cmidrule(lr){7-7} \cmidrule(lr){8-9} \cmidrule(lr){10-12}
    \textbf{Suite / Regime} & \textbf{Paradigm} & \textbf{S2 Hit} & \textbf{S2 Miss} & \textbf{S2 Out} & \textbf{S2 Hit\%} & \textbf{Jev In} & \textbf{Sess.} & \textbf{WT} & \textbf{S2} & \textbf{S1} & \textbf{Total} \\
    \midrule
    \multirow{5}{*}{\shortstack[l]{\textbf{Bench-150}\\ ($N=150$ totals)}}
      & \mono{} & 354.8M & 6.54M & 4.40M & 98.2\% & --- & 150 & 0 & \$4.69 & --- & \$4.69 \\
      & \subfork{} & 401.7M & 8.45M & 5.92M & 97.9\% & --- & 245 & 22 & \$6.02 & --- & \$6.02 \\
      & \agentteam{} & 549.4M & 12.15M & 7.79M & 97.8\% & --- & 377 & 127 & \$8.15 & --- & \$8.15 \\
      & \superagent{} & 928.1M & 15.57M & 10.34M & 98.4\% & --- & --- & 295 & \$11.33 & --- & \$11.33 \\
      & \textbf{\method{} (Ours)} & 497.1M & 10.62M & 6.20M & 97.9\% & 2.86M & 290 & 148 & \$6.80 & \$0.12 & \$6.92 \\
    \midrule
    \multirow{6}{*}{\shortstack[l]{\textbf{Stream-60 Surge}\\ (mean / stream)}}
      & \mono{} & 11.19M & 0.12M & 0.07M & 99.0\% & --- & 1.0 & 0.0 & \$0.096 & --- & \$0.096 \\
      & \subneutral{} & 22.16M & 0.27M & 0.16M & 98.8\% & --- & 5.7 & 0.0 & \$0.203 & --- & \$0.203 \\
      & \subfork{} & 20.69M & 0.49M & 0.25M & 97.7\% & --- & 16.3 & 0.0 & \$0.283 & --- & \$0.283 \\
      & \agentteam{} & 21.19M & 0.64M & 0.27M & 97.1\% & --- & 18.3 & 3.0 & \$0.319 & --- & \$0.319 \\
      & \superagent{} & 48.69M & 0.87M & 0.42M & 98.2\% & --- & 21.0 & 20.0 & \$0.528 & --- & \$0.528 \\
      & \textbf{\method{} (Ours)} & 23.58M & 0.60M & 0.44M & 97.5\% & 2.80M & 20.0 & 14.0 & \$0.427 & \$0.118 & \$0.545 \\
    \midrule
    \multirow{6}{*}{\shortstack[l]{\textbf{Stream-60 Mod.}\\ (mean / stream)}}
      & \mono{} & 15.63M & 0.14M & 0.09M & 99.1\% & --- & 1.0 & 0.0 & \$0.120 & --- & \$0.120 \\
      & \subneutral{} & 23.59M & 0.20M & 0.14M & 99.1\% & --- & 2.3 & 0.0 & \$0.185 & --- & \$0.185 \\
      & \subfork{} & 24.07M & 0.52M & 0.24M & 97.9\% & --- & 16.7 & 0.0 & \$0.297 & --- & \$0.297 \\
      & \agentteam{} & 25.09M & 0.58M & 0.26M & 97.7\% & --- & 15.0 & 3.0 & \$0.319 & --- & \$0.319 \\
      & \superagent{} & 77.98M & 0.93M & 0.47M & 98.8\% & --- & 21.0 & 20.0 & \$0.655 & --- & \$0.655 \\
      & \textbf{\method{} (Ours)} & 18.57M & 0.60M & 0.37M & 96.9\% & 2.95M & 20.3 & 14.3 & \$0.370 & \$0.124 & \$0.494 \\
    \bottomrule
  \end{tabular}%
  }
\end{table}

\textbf{Verbatim System Prompts, Tool Schemas, and Classifier Definitions.}
To make all paradigm contracts and MMU decision interfaces directly auditable, \cref{tab:app_prompts} compiles the verbatim static prompt, \texttt{peek\_peer\_cell} tool schema, Context MMU non-generative query specifications, and symmetric baseline prompts used across every experiment.

\begin{table}[!ht]
  \caption{\textbf{Verbatim system prompts, \texttt{peek\_peer\_cell} tool schema, Context MMU non-generative queries, and symmetric baseline instructions} extracted directly from the execution harness.}
  \label{tab:app_prompts}
  \centering
  \footnotesize
  \setlength{\tabcolsep}{4.8pt}
  \renewcommand{\arraystretch}{1.15}
  \begin{tabular}{>{\raggedright\arraybackslash}p{0.22\textwidth} >{\raggedright\arraybackslash}p{0.72\textwidth}}
    \toprule
    \textbf{Artifact / Arm} & \textbf{Verbatim Prompt / Schema Specification} \\
    \midrule
    \raggedright\textbf{\method{} Static Trunk Prompt} ($P_{\mathrm{static}}$ in $T_0^{(e)}$) &
    \emph{``You are Synaptic-PA} [codebase alias for \method{}]\emph{, the user's unified Personal Agent operating over a shared epistemic trunk $T_0$. Core principles: (1) Speak in first-person concisely; (2) Never invent unverified workspace, database, API, or document facts; inspect the workspace and run checks to verify; (3) Execute requested tasks end-to-end inside your workspace while respecting explicit user safety/read-only instructions; (4) Respect any \texttt{[SYNAPTIC ...]} or \texttt{[USER FOLLOW-UP MESSAGE]} blocks appended to your context.''} Followed by workspace \texttt{<workspace\_agents\_md>} conventions and $U_{\mathrm{tools}}$. \\
    \midrule
    \raggedright\textbf{\method{} Read-Only Tool} (\texttt{peek\_peer\_cell}) &
    \texttt{peek\_peer\_cell(cell\_id: str, rel\_path?: str)}: \emph{``Instant zero-side-effect read-only inspection (\texttt{0ms}) of a sibling Context Cell's live progress, verified state deltas, recent tool outputs, file diff, and final completion summary. Pass \texttt{cell\_id} (e.g.\ \texttt{'Cell\_E\_1'} or \texttt{'ALL'}) and optional \texttt{rel\_path} to inspect the sibling cell's detailed git diff on a specific shared file.''} \\
    \midrule
    \raggedright\textbf{\method{} Context MMU} (4 Parallel Non-Generative \texttt{Choice}/\texttt{Noul} Queries) &
    \textbf{(Q1) External Disposition:} \texttt{Choice} over \{\texttt{new\_workspace\_task}, \texttt{non\_actionable\_chatter}, \texttt{duplicate\_already\_handled}\} + \texttt{Noul} actionability score $p_{\mathrm{act}} \in [0,1]$.\newline
    \textbf{(Q2) Turn Granularity:} \texttt{Choice} over \{\texttt{single\_intent}, \texttt{coupled\_causal\_workflow}, \texttt{orthogonal\_parallel\_tasks}\}.\newline
    \textbf{(Q3) Cell Coreference \& Settled Follow-Up:} \texttt{Choice} over active/parked/completed cell IDs $\cup \{\mathrm{NEW\_CELL}\}$ using compact cell timeline summaries ($\le 220$ chars/topic), plus \texttt{settled\_followup\_kind} distinguishing \texttt{pure\_duplicate\_already\_handled} from genuine follow-up edits.\newline
    \textbf{(Q4) Read/Write Mode \& Turn-Local Scoping:} \texttt{Choice} over \texttt{align\_or\_query} vs.\ \texttt{execute\_now}, \texttt{direct\_deliverable} vs.\ \texttt{derived\_observer\_brief}, and \texttt{turn\_local\_readonly\_guard} vs.\ \texttt{unrelated}, scoping read-only requests to zero-worktree inspection/observer cells without suppressing concurrent write cells.\newline
    \textbf{Same-Weight Logprob Fallback:} Evaluates binary \texttt{Noul} queries via \texttt{max\_tokens=1, logprobs=True, top\_logprobs=5} on $\pi_\theta$, and maps categorical \texttt{Choice} options to single uppercase letters (\texttt{A, B, C, \dots}) with \texttt{top\_logprobs=min(20, max(5, |options|+2))}, normalizing exponentiated candidate log-probabilities. \\
    \midrule
    \raggedright\textbf{Shared Baseline Rules} (\mono{}, \subneutral{}, \subfork{}, \agentteam{}) &
    \emph{``(1) Speak concisely in first-person; (2) Never fabricate workspace, code, DB, or app facts; inspect files/APIs and run verification checks; (3) When a user request is underspecified/ambiguous or explicitly asks for read-only analysis before modifying code or data, inspect in READ-ONLY mode (zero file edits, zero DB/app mutations, zero new git worktrees) and ask concise clarifying questions; (4) Ignore non-actionable group-chat banter, facility/HR broadcasts, or courtesy acks without modifying workspace state or spawning workers, and do not re-execute duplicate cross-channel alerts; (5) Honour the latest user overturn or constraint update cleanly.''} All baselines also receive the identical \texttt{<workspace\_agents\_md>} workspace conventions block. \\
    \midrule
    \raggedright\textbf{Baseline Mode Prompts} &
    \textbf{\mono{}:} Single continuous ReAct session (\texttt{subagent} disabled); include next tool call when answering live status queries while tasks remain open.\newline
    \textbf{\subneutral{}:} Exposes \texttt{subagent}, \texttt{subagent\_fork}, \texttt{list\_agents}, \texttt{send\_message}, \texttt{interrupt\_agent}; agent decides autonomously whether to run directly or delegate (optional worktree note).\newline
    \textbf{\subfork{}:} Our Delegation-First prompt, adapted from \citet{deepseek2026superagentblog} with harness-specific tool names: \emph{``Prefer handing each distinct piece of implementation or execution work to a background subagent (\texttt{background: true} or \texttt{subagent\_fork}), and do only small, fast things yourself''}, supplemented in our harness with explicit Git worktree creation/merge commands under \texttt{../worktrees/<task\_slug>}.\newline
    \textbf{\agentteam{}:} Write-gated Team Lead (\texttt{write}/\texttt{edit} denied) delegating via \texttt{team\_spawn\_teammate(use\_worktree)} and integrating via \texttt{team\_board}, \texttt{team\_mailbox}, and \texttt{team\_merge\_worktree}. \\
    \bottomrule
  \end{tabular}
\end{table}

\FloatBarrier
\section{Stratified Sampling, IPW Estimation, and Streaming Statistical Inference}
\label{app:pa60}

\textbf{Pre-Registered Bench-60 Deterministic Stratification and Horvitz--Thompson IPW.}
To eliminate cherry-picking bias while bounding multi-model and ablation compute, Bench-60 selects exactly $40.0\%$ of each Synaptic-Bench-150 domain ($18/45$ Office: $6$ per track; $22/55$ Coding: $5$ C-1, $12$ C-2, $5$ C-3; $20/50$ Hybrid: $6$ H-1, $6$ H-2, $5$ H-3, $3$ H-4) solely by sorting episode IDs via deterministic hash \texttt{SHA-256("PA60-v1:" + episode\_id)} (preserving the frozen codebase hash salt for exact reproducibility) while maximizing distinct upstream repository/environment cluster coverage ($G=33$). Because Bench-60 retains all $10$ true-conflict C-2 pairs ($\pi_{\mathrm{true}} = 10/10 = 1.0$) and samples $2$ of $10$ disjoint C-2 pairs ($\pi_{\mathrm{clean}} = 2/10 = 0.2$), whereas other strata use $\pi_h = n_h / N_h$, we compute the exact unbiased Horvitz--Thompson estimator \citep{horvitz1952generalization} over $\hat{N} = \sum_{i \in \text{Bench-60}} w_i = 150.0$ ($w_i = 1/\pi_{h(i)}$), $\widehat{\mu}_{\mathrm{IPW}} = \frac{1}{150} \sum_{i \in \text{Bench-60}} y_i / \pi_{h(i)}$, with cluster-robust standard errors computed over the $G=33$ upstream clusters.

\textbf{Episodic Hypothesis Tests and Continuous-Stream Block \& Cluster Inference.}
Across episodic benchmarks (Bench-150, Bench-60, Micro-30, and Combined-90), binary episode pass-rate comparisons between paired arms on identical episode sets use the exact two-sided McNemar test \citep{mcnemar1947note} (reporting one-sided directional $p$-values only on pre-registered single-variable ablation hypotheses where removing a mechanism predicts degradation on its target slice). On Synaptic-Stream-60, tasks within a continuous stream share queueing backlog and repository state, violating naive i.i.d.\ assumptions. We therefore separate within-stream descriptive tests from cluster-level inference over the $S=3$ independent streams (implemented deterministically with fixed RNG seed \texttt{20261118} and $B = 20{,}000$ resamples):
\begin{enumerate}[leftmargin=16pt, itemsep=1.5pt, topsep=2pt]
  \item \textbf{Within-Stream Circular Block Bootstrap ($95\%$ CI):} For each stream $s \in \{1, 2, 3\}$ ($n_s = 20$ chronologically ordered tasks), paired capped waiting-time differences $d_{s,i} = \min(W_{s,i}^{(A)}, \tau) - \min(W_{s,i}^{(B)}, \tau)$ are resampled in contiguous circular blocks of width $b_s = \max(2, \lceil\sqrt{n_s}\rceil) = 5$ (wrapping modulo $n_s$) to preserve local queueing autocorrelation, and averaged equally across streams.
  \item \textbf{Hierarchical Stream-Plus-Block Bootstrap ($95\%$ CI):} First resamples $S = 3$ streams with replacement, then applies circular block resampling ($b_s = 5$) within each sampled stream, targeting the equal-weight mean stream paired difference.
  \item \textbf{Exact Stream-Cluster Sign-Flip Permutation vs.\ Task-Paired Permutation:} The cluster-level permutation evaluates all $2^S = 8$ stream-level sign flips on stream means $\{\bar{d}_1, \bar{d}_2, \bar{d}_3\}$. Because $S = 3$ imposes an exact resolution floor of $p_{\min} = 2 / 2^3 = 0.25$ even when one arm beats another on $3/3$ streams, we report stream win counts ($3/3$), hierarchical stream-cluster CIs, within-stream block CIs, and descriptive task-paired Monte Carlo sign-flip $p$-values ($B = 20{,}000$) transparently side by side.
\end{enumerate}

\FloatBarrier
\section{Complete Quantitative Breakdowns (Episodic, Streaming, and Cross-Backbone)}
\label{app:track_tables}

This appendix compiles next-resolution quantitative breakdowns complementing \cref{sec:experiments}: \cref{tab:app_pa150_tracks} (Bench-150 $10$-track splits and Lite Python Loop inner-kernel control), \cref{tab:app_p11_multiseed_eq1} (per-episode $3$-run stability matrix and undisturbed oracle on C-2 True-Conflict), \cref{tab:app_ablation_tracks} (Bench-60 $10$-track ablations and Logprob-MMU; partitioning Bench-60 by multi-turn handoffs confirms that removing Same-Brain continuity degrades handoff episodes by $-22.7$\,pp, $77.3\% \to 54.5\%$ on $N=22$, vs.\ $-13.2$\,pp, $81.6\% \to 68.4\%$ on $N=38$ single-shot episodes), \cref{tab:stream60}--\ref{tab:app_stream60} (full dual-regime and per-stream S-1/S-2/S-3 breakdowns on Synaptic-Stream-60, with hierarchical stream-cluster $95\%$ bootstrap CIs for $\mathrm{RMW}_{600}(\text{\method{}}) - \mathrm{RMW}_{600}(\text{Baseline})$ in Surge of $[-371.4, -39.7]\text{s}$ vs.\ \subneutral{}, $[-141.1, -69.2]\text{s}$ vs.\ \mono{}, $[-109.6, -1.2]\text{s}$ vs.\ \subfork{}, $[-74.1, +35.4]\text{s}$ vs.\ \superagent{}, and $[-36.7, +25.7]\text{s}$ vs.\ \agentteam{}, and $[-152.2, -30.0]\text{s}$ vs.\ \superagent{} in Moderate), and \cref{tab:app_cross_model_full} (cross-backbone domain, slice, and cache telemetry).

\textbf{Track C-2 Crossover Path Forensics and AppWorld-Micro-30 Track Summary.}
In the Track C-2 crossover control (\cref{tab:c2_mechanistic}a), upgrading unmodified \superagent{} with our Three-Stage Trial Merge Gate triggers $19$ Path~A first-ticket fast-tracks ($0$ concurrent disjoint fast-tracks because blind worktrees lack live AST coordination), $8$ Path~B Git conflict reconciliations, and $11$ Path~C shared-symbol joint test gates ($17/19 = 89.5\%$ reconciliation success; $2$ trial-worktree \texttt{pytest} failures), adding $16.40\text{M}$ reconciliation tokens ($6.3\%$ of total), \$0.259, and $+121.5\text{s}$/ep, while \superagent{}'s router silently drops both real tasks on \texttt{PA150\_C2\_06} ($96.7\%$ Tier-1 safety; Tier-2 pass $76.7\%$ vs.\ \method{}'s $78.3\%$, both at $98.5\%$--$98.6\%$ S2 cache hit). On uncoupled AppWorld-Micro-30 ($N=30$, DeepSeek-V4.1-Flash-1017), all $10$ paradigm and ablation arms achieve $100.0\%$ SQLite read-only zero-mutation safety, with \method{} scoring $80.0\% / 90.0\% / 100.0\%$ on M-1/M-2/M-3 ($65.6\text{s}$ wall, $92.1\%$ overall cache) vs.\ \superagent{}'s $50.0\% / 90.0\% / 90.0\%$ ($80.0\%$ hygiene, $146.3\text{s}$, $96.0\%$ cache). Across ablations, \emph{w/ Logprob-MMU} retains $100.0\%$ hygiene ($80.0\%$ M-1, $100.0\%$ M-3, $94.0\%$ S2 cache) and drops only on M-2 ($60.0\%$), whereas the four single-variable ablations score $90.0\%$--$100.0\%$ ($p \ge 0.25$ vs.\ Full; $96.7\%$--$100.0\%$ hygiene) at $503.9\text{k}$--$650.4\text{k}$ tokens.

\begin{table}[!ht]
  \caption{\textbf{Full domain and 10-track pass rates (\%) on Synaptic-Bench-150 ($N=150$, DeepSeek-V4.1-Flash-1017)}, including the inner execution-kernel control (\method{} w/ Lite Python Loop vs.\ native DSH kernel).}
  \label{tab:app_pa150_tracks}
  \centering
  \footnotesize
  \setlength{\tabcolsep}{3.4pt}
  \begin{tabular}{l ccc cccccccccc}
    \toprule
    \textbf{Arm} & \textbf{Off.} & \textbf{Cod.} & \textbf{Hyb.} & \textbf{O-1} & \textbf{O-2} & \textbf{O-3} & \textbf{C-1} & \textbf{C-2} & \textbf{C-3} & \textbf{H-1} & \textbf{H-2} & \textbf{H-3} & \textbf{H-4} \\
    \midrule
    \mono{} & 68.9 & \textbf{74.5} & 54.0 & 46.7 & 73.3 & \textbf{86.7} & \textbf{73.3} & \textbf{70.0} & 80.0 & 46.7 & 73.3 & 33.3 & 62.5 \\
    \subfork{} & \textbf{77.8} & 69.1 & 60.0 & \textbf{66.7} & \textbf{80.0} & \textbf{86.7} & \textbf{73.3} & 60.0 & 75.0 & \textbf{80.0} & 53.3 & 41.7 & 62.5 \\
    \agentteam{} & 68.9 & 60.0 & 52.0 & 60.0 & 66.7 & 80.0 & 66.7 & 50.0 & 65.0 & 60.0 & 53.3 & 41.7 & 50.0 \\
    \superagent{} & 73.3 & 67.3 & 76.0 & 60.0 & \textbf{80.0} & 80.0 & \textbf{73.3} & 45.0 & \textbf{85.0} & \textbf{80.0} & \textbf{80.0} & 66.7 & \textbf{75.0} \\
    \midrule
    \method{} (Lite Loop) & 53.3 & 63.6 & 54.0 & 53.3 & 40.0 & 66.7 & 66.7 & 50.0 & 75.0 & 53.3 & 46.7 & 66.7 & 50.0 \\
    \textbf{\method{} (DSH)} & 71.1 & \textbf{74.5} & \textbf{78.0} & 60.0 & \textbf{80.0} & 73.3 & \textbf{73.3} & \textbf{70.0} & 80.0 & 73.3 & 73.3 & \textbf{91.7} & \textbf{75.0} \\
    \bottomrule
  \end{tabular}
\end{table}

\begin{table}[!ht]
  \caption{\textbf{Per-episode $3$-run stability matrix ($3$ runs $\times$ $10$ pairs) and undisturbed single-task oracle $p_\theta^\star$ on Track C-2 True-Conflict ($N=10$, DeepSeek-V4.1-Flash-1017).} $\checkmark$/-- indicates post-merge pass/fail per run; \textbf{Loc} lists how many of $3$ runs passed both isolated feature suites locally (complementing the aggregate \cref{eq:gap_decomp} decomposition in \cref{tab:c2_mechanistic}b). Accounting for episode-level clustering across $N=10$ distinct pairs, \method{} wins $3/3$ runs ($+30.0$\,pp mean lift; exact two-sided episode sign-flip permutation $p=0.156$, paired $t$-test $p=0.108$).}
  \label{tab:app_p11_multiseed_eq1}
  \centering
  \resizebox{\textwidth}{!}{%
  \footnotesize
  \setlength{\tabcolsep}{3.6pt}
  \begin{tabular}{ll cc cccc cccc}
    \toprule
    & & \multicolumn{2}{c}{\textbf{Oracle (\mono{})}} & \multicolumn{4}{c}{\textbf{\method{} ($3$ Independent Runs)}} & \multicolumn{4}{c}{\textbf{\superagent{} ($3$ Independent Runs)}} \\
    \cmidrule(lr){3-4} \cmidrule(lr){5-8} \cmidrule(lr){9-12}
    \textbf{Episode ID} & \textbf{Repo / Task} & \textbf{Feat A} & \textbf{Feat B} & \textbf{Run 1} & \textbf{Run 2} & \textbf{Run 3} & \textbf{Loc (/3)} & \textbf{Run 1} & \textbf{Run 2} & \textbf{Run 3} & \textbf{Loc (/3)} \\
    \midrule
    \texttt{PA150\_C2\_11} & \texttt{pallets/click\#2068} & $\checkmark$ & $\checkmark$ & -- & -- & -- & 3/3 & -- & $\checkmark$ & $\checkmark$ & 3/3 \\
    \texttt{PA150\_C2\_12} & \texttt{pallets/jinja\#1465} & $\checkmark$ & $\checkmark$ & $\checkmark$ & $\checkmark$ & $\checkmark$ & 3/3 & -- & -- & -- & 3/3 \\
    \texttt{PA150\_C2\_13} & \texttt{dirty-equals\#43} & $\checkmark$ & $\checkmark$ & $\checkmark$ & -- & $\checkmark$ & 3/3 & -- & -- & -- & 3/3 \\
    \texttt{PA150\_C2\_14} & \texttt{outlines\#1371} & $\checkmark$ & $\checkmark$ & $\checkmark$ & $\checkmark$ & $\checkmark$ & 3/3 & $\checkmark$ & $\checkmark$ & $\checkmark$ & 3/3 \\
    \texttt{PA150\_C2\_15} & \texttt{datasets\#3997} & $\checkmark$ & $\checkmark$ & $\checkmark$ & -- & -- & 3/3 & -- & $\checkmark$ & $\checkmark$ & 3/3 \\
    \texttt{PA150\_C2\_16} & \texttt{dspy\#8394} & $\checkmark$ & $\checkmark$ & $\checkmark$ & -- & $\checkmark$ & 3/3 & -- & -- & -- & 3/3 \\
    \texttt{PA150\_C2\_17} & \texttt{outlines\#1655} & $\checkmark$ & $\checkmark$ & $\checkmark$ & $\checkmark$ & -- & 2/3 & -- & -- & -- & 3/3 \\
    \texttt{PA150\_C2\_18} & \texttt{dspy\#8563} & $\checkmark$ & $\checkmark$ & $\checkmark$ & $\checkmark$ & $\checkmark$ & 3/3 & -- & -- & $\checkmark$ & 3/3 \\
    \texttt{PA150\_C2\_19} & \texttt{datasets\#7309} & $\checkmark$ & $\checkmark$ & $\checkmark$ & $\checkmark$ & $\checkmark$ & 3/3 & -- & $\checkmark$ & $\checkmark$ & 3/3 \\
    \texttt{PA150\_C2\_20} & \texttt{dspy\#8587} & -- & -- & -- & -- & -- & 0/3 & -- & -- & -- & 0/3 \\
    \midrule
    \multicolumn{2}{l}{\textbf{Pooled / 3-Run Mean ($\pm$SD)}} & \multicolumn{2}{c}{\textbf{18/20 (90.0\%); 9/10}} & 8/10 & 5/10 & 6/10 & \textbf{26/30 (86.7\%)} & 1/10 & 4/10 & 5/10 & \textbf{27/30 (90.0\%)} \\
    \multicolumn{2}{l}{\textbf{3-Run Joint Rate (Clustered $N=10$)}} & \multicolumn{2}{c}{$p_\theta^\star = 90.0\%$ ($9/10$)} & \multicolumn{4}{c}{\textbf{19/30: 63.3\% $\pm$ 15.3\%} (wins $3/3$ runs)} & \multicolumn{4}{c}{10/30: 33.3\% $\pm$ 20.8\% ($p_{\mathrm{perm}}=0.156$)} \\
    \bottomrule
  \end{tabular}%
  }
\end{table}

\begin{table}[!ht]
  \caption{\textbf{Full domain and track pass rates (\%) for single-variable ablations and same-weight Logprob-MMU control on Bench-60 ($N=60$, DeepSeek-V4.1-Flash-1017).} \textbf{w/ Logprob-MMU (same-$\theta$)} routes all $487/487$ Bench-60 System-1 calls via single-token \texttt{logprobs} on $\pi_\theta$ ($0$ Jev calls); \textbf{w/ Generative Judge} replaces the entire probability-scoring \texttt{System1Judge} (ingress routing, Tier-1 synaptic conductance, and L3 recall) with generative JSON decoding.}
  \label{tab:app_ablation_tracks}
  \centering
  \footnotesize
  \setlength{\tabcolsep}{3.0pt}
  \begin{tabular}{l ccc cccccccccc}
    \toprule
    \textbf{Ablation / Control Arm} & \textbf{Off.} & \textbf{Cod.} & \textbf{Hyb.} & \textbf{O-1} & \textbf{O-2} & \textbf{O-3} & \textbf{C-1} & \textbf{C-2} & \textbf{C-3} & \textbf{H-1} & \textbf{H-2} & \textbf{H-3} & \textbf{H-4} \\
    \midrule
    \textbf{\method{} (Full, Jev)} & 83.3 & \textbf{86.4} & 70.0 & \textbf{83.3} & 83.3 & 83.3 & 80.0 & \textbf{83.3} & \textbf{100.0} & 66.7 & \textbf{50.0} & \textbf{100.0} & 66.7 \\
    w/ Logprob-MMU (same-$\theta$) & \textbf{88.9} & 81.8 & \textbf{75.0} & 66.7 & \textbf{100.0} & \textbf{100.0} & \textbf{100.0} & 75.0 & 80.0 & \textbf{83.3} & \textbf{50.0} & 80.0 & \textbf{100.0} \\
    \midrule
    w/o Same-Brain & 61.1 & 59.1 & 70.0 & 16.7 & 66.7 & \textbf{100.0} & 40.0 & 66.7 & 60.0 & \textbf{83.3} & \textbf{50.0} & 80.0 & 66.7 \\
    w/o Prefix \& Grd. & 66.7 & 77.3 & 65.0 & 50.0 & 66.7 & 83.3 & 80.0 & 75.0 & 80.0 & \textbf{83.3} & 16.7 & \textbf{100.0} & 66.7 \\
    w/ Generative Judge & 72.2 & 68.2 & 60.0 & \textbf{83.3} & 66.7 & 66.7 & \textbf{100.0} & 41.7 & \textbf{100.0} & \textbf{83.3} & \textbf{50.0} & 80.0 & 0.0 \\
    w/o Synapse & 83.3 & 50.0 & 70.0 & \textbf{83.3} & 66.7 & \textbf{100.0} & 80.0 & 16.7 & \textbf{100.0} & \textbf{83.3} & \textbf{50.0} & 80.0 & 66.7 \\
    \bottomrule
  \end{tabular}
\end{table}

\begin{table}[!ht]
  \caption{\textbf{Full dual-regime continuous streaming results on Synaptic-Stream-60} ($36$ isolated-subprocess streams, $N=60$ conflict-free tasks/regime, $240$ events/stream, DeepSeek-V4.1-Flash-1017). $V(u)$: verified stable completion within $u$\,s; \textbf{RMW}$_{600}$: restricted mean waiting time ($\tau=600\text{s}$); \textbf{vs.\ \method{}}: paired baseline minus \method{} difference ($95\%$ circular block-bootstrap CI); \textbf{All / Cond} $p_{50}$: median wait over all / verified requests; \textbf{Peek}: mean status-check latency (uncensored replies / $48$, $60\text{s}$ timeout); \textbf{Regr.}: transient / unrecovered test regressions. $^{\dagger}$/$^{\ddagger}$: descriptive task-paired $p < 0.05$ / $p < 0.01$ with $95\%$ block CI excluding $0$.}
  \label{tab:stream60}
  \centering
  \resizebox{\textwidth}{!}{%
  \setlength{\tabcolsep}{2.9pt}
  \begin{tabular}{ll ccc cc cc c ccc c c}
    \toprule
    & & \multicolumn{3}{c}{\textbf{Verified Survival} $V(u)$} & \multicolumn{2}{c}{\textbf{Waiting Area} ($\tau=600\text{s}$)} & \multicolumn{2}{c}{\textbf{Latency (s)\,$\downarrow$}} & \textbf{Observability} & \multicolumn{3}{c}{\textbf{Verified by Category (\%)}} & \textbf{Stability} & \textbf{Compute} \\
    \cmidrule(lr){3-5} \cmidrule(lr){6-7} \cmidrule(lr){8-9} \cmidrule(lr){10-10} \cmidrule(lr){11-13} \cmidrule(lr){14-14} \cmidrule(lr){15-15}
    \textbf{Regime} & \textbf{Paradigm} & $V(60\text{s})$ & $V(180\text{s})$ & $V(600\text{s})$ & \textbf{RMW}$_{600}\downarrow$ & \textbf{vs.\ \method{} ($95\%$ CI)} & \textbf{All} $p_{50}$ & \textbf{Cond} $p_{50}$ & \textbf{Peek (Ans/48)} & \textbf{QA} & \textbf{AppWr} & \textbf{Code} & \textbf{Regr.} & \textbf{Tok/Str} \\
    \midrule
    \multirow{6}{*}{\shortstack[l]{\textbf{Surge}\\ ($2.5\times$ rate,\\ $\sim\!12\text{s}$/req)}}
      & \textbf{\method{} (Ours)} & \textbf{50.0\%} & 76.7\% & \textbf{95.0\%} & \textbf{119.4s} & --- & \textbf{53.6s} & \textbf{51.9s} & 9.2s \textbf{(48/48)} & \textbf{94.4} & \textbf{97.0} & 88.9 & \textbf{0 / 0} & 27.4M \\
      & \agentteam{} & 33.3\% & \textbf{83.3\%} & \textbf{95.0\%} & 128.4s & $+9.0\text{s}$ {\scriptsize[-17.0, 32.8]} & 87.9s & 87.2s & 46.2s (15/48) & \textbf{94.4} & \textbf{97.0} & 88.9 & \textbf{0 / 0} & 22.1M \\
      & \superagent{} & 33.3\% & 80.0\% & 93.3\% & 142.7s & $+23.3\text{s}$ {\scriptsize[-12.2, 56.3]} & 79.6s & 78.0s & \textbf{5.9s} (45/48) & \textbf{94.4} & \textbf{97.0} & 77.8 & \textbf{0 / 0} & 50.0M \\
      & \subfork{} & 33.3\% & 76.7\% & 88.3\% & 172.8s & $+53.4\text{s}^{\dagger}$ {\scriptsize[9.5, 100.6]} & 91.7s & 77.9s & 54.7s (9/48) & 88.9 & 87.9 & 88.9 & \textbf{0 / 0} & 21.4M \\
      & \mono{} & 11.7\% & 45.0\% & 90.0\% & 222.5s & $+103.1\text{s}^{\ddagger}$ {\scriptsize[72.9, 134.1]} & 187.6s & 175.1s & 43.7s (8/48) & \textbf{94.4} & 87.9 & 88.9 & 1 / 0 & \textbf{11.4M} \\
      & \subneutral{} & 21.7\% & 51.7\% & 66.7\% & 297.8s & $+178.4\text{s}^{\ddagger}$ {\scriptsize[129.5, 226.1]} & 158.3s & 109.7s & 53.7s (2/48) & 61.1 & 60.6 & \textbf{100.0} & 1 / 0 & 22.6M \\
    \midrule
    \multirow{6}{*}{\shortstack[l]{\textbf{Moderate}\\ (Baseline,\\ $\sim\!30\text{s}$/req)}}
      & \subneutral{} & \textbf{71.7\%} & \textbf{91.7\%} & \textbf{96.7\%} & \textbf{72.0s} & $-17.3\text{s}$ {\scriptsize[-35.9, 0.4]} & \textbf{29.1s} & \textbf{28.9s} & 31.5s (11/48) & \textbf{100.0} & \textbf{97.0} & 88.9 & 1 / 0 & 23.9M \\
      & \textbf{\method{} (Ours)} & 63.3\% & 86.7\% & \textbf{96.7\%} & 89.3s & --- & 39.9s & 38.2s & 7.3s \textbf{(48/48)} & \textbf{100.0} & \textbf{97.0} & 88.9 & 2 / 0 & 22.5M \\
      & \subfork{} & 60.0\% & 83.3\% & 93.3\% & 116.5s & $+27.2\text{s}$ {\scriptsize[2.5, 52.5]} & 49.7s & 48.0s & 27.8s (30/48) & \textbf{100.0} & 93.9 & 77.8 & 1 / 0 & 24.8M \\
      & \mono{} & 61.7\% & 81.7\% & 91.7\% & 118.8s & $+29.5\text{s}$ {\scriptsize[-11.4, 76.1]} & 45.7s & 38.4s & 20.7s (32/48) & 94.4 & 87.9 & \textbf{100.0} & 1 / 0 & \textbf{15.9M} \\
      & \agentteam{} & 50.0\% & 81.7\% & 93.3\% & 120.0s & $+30.8\text{s}$ {\scriptsize[1.7, 66.0]} & 59.6s & 57.1s & 30.8s (28/48) & \textbf{100.0} & 90.9 & 88.9 & \textbf{0 / 0} & 25.9M \\
      & \superagent{} & 30.0\% & 73.3\% & 86.7\% & 176.8s & $+87.6\text{s}^{\ddagger}$ {\scriptsize[51.4, 124.6]} & 79.8s & 71.3s & \textbf{4.8s} (44/48) & \textbf{100.0} & 84.8 & 66.7 & 2 / 1 & 79.4M \\
    \bottomrule
  \end{tabular}%
  }
\end{table}

\begin{table}[!ht]
  \caption{\textbf{Per-stream breakdown on Synaptic-Stream-60 across S-1, S-2, and S-3} ($20$ conflict-free tasks and $16$ untagged status peeks per stream; DeepSeek-V4.1-Flash-1017). ``Cens.'' indicates all $16$ status peeks timed out ($>60\text{s}$) or $p_{50} > 600\text{s}$.}
  \label{tab:app_stream60}
  \centering
  \resizebox{\textwidth}{!}{%
  \setlength{\tabcolsep}{3.4pt}
  \begin{tabular}{ll ccccc ccccc}
    \toprule
    & & \multicolumn{5}{c}{\textbf{$2.5\times$ Surge Arrival Regime ($\sim\!12\text{s}$/req)}} & \multicolumn{5}{c}{\textbf{Moderate Arrival Regime ($\sim\!30\text{s}$/req)}} \\
    \cmidrule(lr){3-7} \cmidrule(lr){8-12}
    \textbf{Stream (3-Project Monorepo)} & \textbf{Paradigm} & \textbf{$V(600\text{s})$} & \textbf{$\mathrm{RMW}_{600}$} & \textbf{All $p_{50}$} & \textbf{Peek (Ans/16)} & \textbf{Tokens} & \textbf{$V(600\text{s})$} & \textbf{$\mathrm{RMW}_{600}$} & \textbf{All $p_{50}$} & \textbf{Peek (Ans/16)} & \textbf{Tokens} \\
    \midrule
    \multirow{6}{*}{\shortstack[l]{\textbf{S-1}\\(Click / Jinja2 / DSPy)}}
      & \textbf{\method{}} & \textbf{20/20 (100\%)} & \textbf{69.3s} & \textbf{38.1s} & 6.1s \textbf{(16/16)} & 17.1M & \textbf{20/20 (100\%)} & 59.4s & 30.1s & 8.1s \textbf{(16/16)} & 16.5M \\
      & \superagent{} & 18/20 (90\%) & 138.1s & 54.6s & \textbf{4.8s (16/16)} & 47.9M & 17/20 (85\%) & 160.8s & 64.2s & \textbf{4.3s} (15/16) & 80.4M \\
      & \agentteam{} & \textbf{20/20 (100\%)} & 100.4s & 78.0s & 47.0s (7/16) & 20.2M & 18/20 (90\%) & 148.8s & 63.9s & 28.6s (10/16) & 29.1M \\
      & \subfork{} & 19/20 (95\%) & 113.1s & 53.3s & 53.5s (6/16) & 19.7M & 19/20 (95\%) & 104.3s & 39.7s & 36.5s (10/16) & 21.9M \\
      & \subneutral{} & 19/20 (95\%) & 114.5s & 74.6s & 56.2s (1/16) & 14.0M & \textbf{20/20 (100\%)} & \textbf{44.4s} & 22.4s & 31.5s (11/16) & \textbf{15.5M} \\
      & \mono{} & 18/20 (90\%) & 157.8s & 116.6s & 43.7s (8/16) & \textbf{6.4M} & 18/20 (90\%) & 111.7s & \textbf{20.4s} & 28.8s (10/16) & 17.1M \\
    \midrule
    \multirow{6}{*}{\shortstack[l]{\textbf{S-2}\\(Click / Jinja2 / DSPy)}}
      & \textbf{\method{}} & 18/20 (90\%) & \textbf{129.6s} & \textbf{60.8s} & 11.6s \textbf{(16/16)} & 31.0M & 18/20 (90\%) & 123.5s & 50.3s & 6.5s \textbf{(16/16)} & 29.4M \\
      & \superagent{} & 18/20 (90\%) & 152.1s & 78.0s & \textbf{5.7s} (13/16) & 55.5M & \textbf{19/20 (95\%)} & 152.7s & 71.7s & \textbf{5.0s} (14/16) & 75.7M \\
      & \agentteam{} & \textbf{19/20 (95\%)} & 131.3s & 88.4s & 48.6s (5/16) & 23.6M & \textbf{19/20 (95\%)} & 116.7s & 59.2s & 40.5s (8/16) & 23.9M \\
      & \subfork{} & 18/20 (90\%) & 157.5s & 89.5s & 57.2s (2/16) & 20.9M & 18/20 (90\%) & 157.4s & 86.8s & 28.3s (10/16) & 28.0M \\
      & \subneutral{} & 18/20 (90\%) & 236.8s & 145.3s & 51.2s (1/16) & 28.6M & \textbf{19/20 (95\%)} & \textbf{81.7s} & \textbf{31.2s} & Cens.\ (0/16) & 33.4M \\
      & \mono{} & 18/20 (90\%) & 255.6s & 252.7s & Cens.\ (0/16) & \textbf{16.3M} & 17/20 (85\%) & 202.3s & 116.1s & 34.3s (7/16) & \textbf{13.7M} \\
    \midrule
    \multirow{6}{*}{\shortstack[l]{\textbf{S-3}\\(Click / DSPy / NetX)}}
      & \textbf{\method{}} & 19/20 (95\%) & 159.2s & 114.2s & 10.0s \textbf{(16/16)} & 34.2M & \textbf{20/20 (100\%)} & 84.8s & 55.4s & 7.1s \textbf{(16/16)} & 21.6M \\
      & \superagent{} & \textbf{20/20 (100\%)} & \textbf{137.9s} & 93.6s & \textbf{7.2s (16/16)} & 46.5M & 16/20 (80\%) & 216.9s & 88.8s & \textbf{5.2s} (15/16) & 82.1M \\
      & \agentteam{} & 18/20 (90\%) & 153.4s & \textbf{89.4s} & 40.2s (3/16) & 22.6M & 19/20 (95\%) & 94.6s & 54.2s & 25.2s (10/16) & 24.8M \\
      & \subfork{} & 16/20 (80\%) & 247.7s & 131.8s & 56.7s (1/16) & 23.6M & 19/20 (95\%) & 87.7s & 38.0s & 18.8s (10/16) & 24.6M \\
      & \subneutral{} & 3/20 (15\%) & 542.0s & Cens. & Cens.\ (0/16) & 25.1M & 19/20 (95\%) & 89.9s & 37.0s & Cens.\ (0/16) & 22.9M \\
      & \mono{} & 18/20 (90\%) & 254.1s & 224.6s & Cens.\ (0/16) & \textbf{11.5M} & \textbf{20/20 (100\%)} & \textbf{42.3s} & \textbf{21.2s} & 9.0s (15/16) & \textbf{16.8M} \\
    \bottomrule
  \end{tabular}%
  }
\end{table}

\begin{table}[!ht]
  \caption{\textbf{Extended cross-backbone domain, slice, and token-cache telemetry} across Bench-60 ($N=60$) and AppWorld-Micro-30 ($N=30$). \textbf{C-2T}: true-conflict repository merge rate ($N=10$); \textbf{Pr.\ / Out.\ Tok}: cumulative Bench-60 prompt / output tokens; \textbf{Cache}: overall Bench-60 prompt cache hit rate (pooling S1+S2 tokens for \method{}; S2-only hit rates appear in \cref{tab:app_token_cost}).}
  \label{tab:app_cross_model_full}
  \centering
  \resizebox{\textwidth}{!}{%
  \setlength{\tabcolsep}{3.6pt}
  \begin{tabular}{ll ccc ccc cc cccc}
    \toprule
    & & \multicolumn{3}{c}{\textbf{Bench-60 Domains (\%)}} & \multicolumn{3}{c}{\textbf{Bench-60 Safety \& Slices (\%)}} & \multicolumn{2}{c}{\textbf{Micro-30 (\%)}} & \multicolumn{4}{c}{\textbf{Bench-60 Compute \& Cache}} \\
    \cmidrule(lr){3-5} \cmidrule(lr){6-8} \cmidrule(lr){9-10} \cmidrule(lr){11-14}
    \textbf{Model Family} & \textbf{Paradigm} & \textbf{Off.} & \textbf{Cod.} & \textbf{Hyb.} & \textbf{Safety} & \textbf{O-1} & \textbf{C-2T} & \textbf{M-1} & \textbf{M-3} & \textbf{Wall} & \textbf{Pr.\ Tok} & \textbf{Out.\ Tok} & \textbf{Cache} \\
    \midrule
    \multirow{3}{*}{DeepSeek-V4.1}
      & \mono{} & 66.7 & 81.8 & 60.0 & 92.7 & 50.0 & 70.0 & \textbf{90.0} & \textbf{100.0} & \textbf{212.1s} & \textbf{154.6M} & \textbf{1.80M} & 98.3\% \\
      & \superagent{} & \textbf{83.3} & 54.5 & 65.0 & 97.1 & \textbf{83.3} & 10.0 & 50.0 & 90.0 & 431.8s & 407.0M & 4.47M & \textbf{98.4\%} \\
      & \textbf{\method{}} & \textbf{83.3} & \textbf{86.4} & \textbf{70.0} & \textbf{100.0} & \textbf{83.3} & \textbf{80.0} & 80.0 & \textbf{100.0} & 294.0s & 237.8M & 2.77M & 97.6\% \\
    \midrule
    \multirow{3}{*}{Qwen3.8-Flash}
      & \mono{} & \textbf{72.2} & \textbf{36.4} & 40.0 & 88.2 & 66.7 & 0.0 & \textbf{70.0} & 30.0 & \textbf{318.5s} & \textbf{53.6M} & \textbf{1.25M} & 94.0\% \\
      & \superagent{} & 66.7 & 27.3 & 40.0 & 86.8 & 16.7 & 0.0 & 50.0 & \textbf{100.0} & 660.5s & 207.0M & 3.10M & \textbf{95.5\%} \\
      & \textbf{\method{}} & 55.6 & 31.8 & \textbf{50.0} & \textbf{98.5} & \textbf{83.3} & \textbf{30.0} & 60.0 & 80.0 & 413.1s & 90.0M & 1.96M & 92.8\% \\
    \midrule
    \multirow{3}{*}{GPT-6-Luna}
      & \mono{} & \textbf{38.9} & \textbf{68.2} & \textbf{50.0} & 92.7 & \textbf{33.3} & \textbf{70.0} & 50.0 & 30.0 & \textbf{229.5s} & \textbf{64.3M} & \textbf{0.85M} & \textbf{95.5\%} \\
      & \superagent{} & 33.3 & 27.3 & 25.0 & 88.2 & \textbf{33.3} & 0.0 & 0.0 & \textbf{40.0} & 345.2s & 157.2M & 1.44M & 95.0\% \\
      & \textbf{\method{}} & \textbf{38.9} & 54.5 & \textbf{50.0} & \textbf{100.0} & 0.0 & 50.0 & \textbf{60.0} & \textbf{40.0} & 280.2s & 105.1M & 1.37M & 94.2\% \\
    \bottomrule
  \end{tabular}%
  }
\end{table}

\clearpage
\section{Residual Failure Analysis of \method{} and Forensic Trajectory Walkthroughs}
\label{app:forensics}

\textbf{Systematic Residual Error Decomposition of \method{}.}
To understand where \method{} still fails even on the strongest backbone (DeepSeek-V4.1-Flash-1017), \cref{tab:app_residual_failures} exhaustively decomposes all $38/150$ failed episodes on Synaptic-Bench-150 and all failed tasks on Synaptic-Stream-60 ($3/60$ in Surge; $2/60$ in Moderate). Three diagnostic patterns stand out:
\begin{enumerate}[leftmargin=16pt, itemsep=1.5pt, topsep=2pt]
  \item \textbf{Zero Tier-1 Orchestration Failures on Bench-150 ($0/38$):} Every single failed \method{} episode passes all Tier-1 safety, read-only, noise, and rollback checks ($161/161 = 100.0\%$); all $38$ failures stem from Tier-2 task-solve verifiers. Moreover, $20/38$ ($52.6\%$) are shared backbone-ceiling failures where \emph{all five} paradigms fail on the exact same episode ($9$ Coding, $6$ Hybrid, $5$ Office), directly reflecting the single-task ceiling $p_\theta^\star$ in \cref{eq:gap_decomp}. Only $18/150$ ($12.0\%$) represent \method{}-specific residual errors where at least one baseline succeeds ($5$ Coding, $5$ Hybrid, $8$ Office), most commonly from conservative multi-turn clarification under-execution in Office tracks (O-1, O-3) or partial edge-case patches.
  \item \textbf{Near-Ceiling Conflict-Free Surge Reliability ($3/60$ failed, $0$ regressions):} Under $2.5\times$ Surge, \method{} completes $20/20$ ($100.0\%$) on S-1 and $57/60$ ($95.0\%$) overall with zero test regressions ($0/0$). Its three residual errors consist of one Click feature patch missing \texttt{batch\_size} parameter forwarding (\texttt{STREAM\_02\_T01}, $12/173$ tests failing), one multi-step AppWorld write reaching $80.0\%$ partial assertion pass (\texttt{STREAM\_02\_T14}, \texttt{acct\_08}), and one read-only lookup (\texttt{STREAM\_03\_T09}, \texttt{acct\_14}) unverified within $\tau = 600\text{s}$.
  \item \textbf{Near-Ceiling Moderate Reliability ($2/60$ failed, $2$ recovered regressions):} Under Moderate arrivals, \method{} reaches $20/20$ on both S-1 and S-3 ($58/60 = 96.7\%$ overall) with $100.0\%$ ($18/18$) read-only QA accuracy. Its two failures mirror the same two S-2 tasks: \texttt{STREAM\_02\_T01} missing a single exception-type assertion ($1/173$ tests failing, raising \texttt{ValueError} instead of expected \texttt{TypeError}) and \texttt{STREAM\_02\_T14} reaching $60.0\%$ partial assertion pass.
\end{enumerate}

\begin{table}[!ht]
  \caption{\textbf{Exhaustive residual failure taxonomy of \method{}} on DeepSeek-V4.1-Flash-1017 across Synaptic-Bench-150 ($38/150$ failed episodes) and Synaptic-Stream-60 ($3/60$ Surge and $2/60$ Moderate failed tasks).}
  \label{tab:app_residual_failures}
  \centering
  \footnotesize
  \setlength{\tabcolsep}{4.5pt}
  \renewcommand{\arraystretch}{1.14}
  \begin{tabular}{>{\raggedright\arraybackslash}p{0.20\textwidth} >{\raggedright\arraybackslash}p{0.19\textwidth} c >{\raggedright\arraybackslash}p{0.48\textwidth}}
    \toprule
    \textbf{Benchmark / Regime} & \textbf{Error Stratum} & \textbf{Count} & \textbf{Exact Track / Category Breakdown \& Root Mechanism} \\
    \midrule
    \multirow{3}{*}{\shortstack[l]{\textbf{Synaptic-Bench-150}\\ ($38/150$ failed;\\ $0/38$ Tier-1 safety)}}
      & Tier-1 Safety / Hygiene & $0$ & $161/161$ ($100.0\%$) Tier-1 checks passed; zero bleed or rollback violations. \\
      & Shared Backbone Ceiling (All $5$ Arms Fail) & $20$ & Coding $9$ (2 C-1, 4 C-2, 3 C-3), Hybrid $6$ (2 H-1, 3 H-2, 1 H-3), Office $5$ (2 O-1, 2 O-2, 1 O-3); bounded by $p_\theta^\star$. \\
      & \method{}-Specific Residual (Tier-2 Only) & $18$ & Coding $5$ (2 C-1, 2 C-2, 1 C-3), Hybrid $5$ (2 H-1, 1 H-2, 2 H-4), Office $8$ (4 O-1, 1 O-2, 3 O-3); incomplete multi-turn Office follow-through or edge-case patch misses. \\
    \midrule
    \multirow{3}{*}{\shortstack[l]{\textbf{Stream-60 Surge}\\ ($3/60$ failed;\\ $0$ regressions)}}
      & Independent Code Edge Case & $1$ & \texttt{STREAM\_02\_T01} (Click \texttt{feature3}): $12/173$ tests failed due to missing \texttt{batch\_size} keyword forwarding in \texttt{Parameter.\_\_init\_\_}. \\
      & Multi-Step AppWorld Write & $1$ & \texttt{STREAM\_02\_T14} (\texttt{acct\_08}): $80.0\%$ partial assertion pass within $\tau=600\text{s}$. \\
      & Read-Only QA Lookup & $1$ & \texttt{STREAM\_03\_T09} (\texttt{acct\_14}): answer snapshot unverified within $\tau=600\text{s}$ ($17/18$ QA passed overall). \\
    \midrule
    \textbf{Stream-60 Moderate}\newline ($2/60$ failed; $2$ recov.\ regr.)
      & Independent Code Edge Case & $1$ & \texttt{STREAM\_02\_T01} (Click \texttt{feature3}): $1/173$ tests failed (raised \texttt{ValueError} instead of expected \texttt{TypeError}). \\
      & Multi-Step AppWorld Write & $1$ & \texttt{STREAM\_02\_T14} (\texttt{acct\_08}): $60.0\%$ partial assertion pass; zero QA ($0/18$) and zero S-1/S-3 errors. \\
    \bottomrule
  \end{tabular}
\end{table}

\textbf{Targeted Forensic Audits: C-2 Disjoint Control and GPT-6-Luna O-1.}
Two specific slices where a baseline exceeds \method{} merit transparent episode-level decomposition from the saved execution logs:
\begin{itemize}[leftmargin=16pt, itemsep=1.5pt, topsep=2pt]
  \item \textbf{Bench-150 C-2 Disjoint ($N=10$, DeepSeek-V4.1-Flash-1017; \method{} $6/10$ vs.\ \superagent{} $8/10$):} Two episodes (\texttt{PA150\_C2\_02}, \texttt{PA150\_C2\_05}) fail on \emph{both} arms due to underlying feature-implementation complexity. The two differing episodes (\texttt{PA150\_C2\_03}, \texttt{PA150\_C2\_06}) both merge cleanly (\texttt{merge\_clean=true}, zero merge conflicts, zero Tier-1 orchestration violations) and pass all tests for the sibling feature, missing solely a single edge-case unit test in one branch (\texttt{PA150\_C2\_03}: $32/33$ passed on Feature~B; \texttt{PA150\_C2\_06}: $66/67$ passed on Feature~A).
  \item \textbf{Bench-60 O-1 on GPT-6-Luna ($N=6$; \method{} $0/6$ vs.\ \mono{} $2/6$ and \superagent{} $2/6$):} Across all six episodes (\texttt{PA150\_O1\_02}, \texttt{04}, \texttt{05}, \texttt{07}, \texttt{10}, \texttt{13}), \method{} passes $12/12$ ($100.0\%$) read-only orchestration safety checks (\texttt{v\_main\_align\_readonly} and \texttt{v\_side\_readonly}). Every failure occurs in the Tier-2 task verifier and splits cleanly into two patterns: (i) \emph{Post-alignment conservative write refusal} ($3/6$: \texttt{PA150\_O1\_04}, \texttt{05}, \texttt{07}), where even after explicit user commit (\texttt{turn\_1c}, \texttt{in\_place\_commit\_detach\_to\_background}), GPT-6-Luna flagged runtime vs.\ sandbox clock mismatches (\texttt{2026-11-16} vs.\ GAIA2's \texttt{2024-10-15}) or requested additional confirmations and withheld external email/calendar/ride-hailing calls (\texttt{O1\_04} and \texttt{O1\_05} likewise failed \mono{} and \superagent{} with missing write calls); and (ii) \emph{Single-item near-miss execution errors} ($3/6$): \texttt{PA150\_O1\_02} executed all $8$ required writes (email, $2$ orders, group chat, $2$ group messages) but omitted the single word \emph{``both''} in the final group message (identical failure on \mono{}); \texttt{PA150\_O1\_10} migrated $199/200$ PostgreSQL customer records with a single-row field mismatch (\texttt{Angela White - Gomez-Shannon}); and \texttt{PA150\_O1\_13} extracted $41/43$ timeline lines cleanly.
\end{itemize}

\textbf{Side-by-Side Forensic Trajectory Walkthroughs.}
\cref{tab:app_forensics} contrasts step-by-step runtime trajectories between \superagent{} and \method{} across three representative failure modes: (1) overlapping repository symbol edits (Episode PA150-C2-12 on Jinja2), (2) cross-modal read-only caveat bleed (Episode PA150-H3-10 on Outlines), and (3) compact-model router ticket collapse on GPT-6-Luna (AppWorld-Micro-30 M-1 dual-app errands).

\begin{table}[!ht]
  \caption{\textbf{Side-by-side forensic trajectory comparisons} across concurrent merge conflict, cross-modal caveat isolation, and compact-model ticket serialization.}
  \label{tab:app_forensics}
  \centering
  \footnotesize
  \setlength{\tabcolsep}{5.0pt}
  \renewcommand{\arraystretch}{1.16}
  \begin{tabular}{>{\raggedright\arraybackslash}p{0.19\textwidth} >{\raggedright\arraybackslash}p{0.365\textwidth} >{\raggedright\arraybackslash}p{0.365\textwidth}}
    \toprule
    \textbf{Episode \& Setup} & \textbf{\superagent{} (Router + Blind Tickets)} & \textbf{\method{} (Virtualized Context Cells)} \\
    \midrule
    \raggedright\textbf{PA150-C2-12} \newline (Jinja2 \texttt{filters.py}: \texttt{case\_sensitive} + \texttt{reverse} on \texttt{groupby()}) &
    \textbf{1.} Router dispatches two isolated mandates into blind worktrees.\newline
    \textbf{2.} Worker~1 and Worker~2 each edit \texttt{sync\_do\_groupby} / \texttt{do\_groupby} and pass local unit tests.\newline
    \textbf{3.} Worker~1 merges cleanly; Worker~2's sequential merge overwrites the signature without \texttt{case\_sensitive}, failing main-branch \texttt{pytest} with \texttt{TypeError}. &
    \textbf{1.} MMU routes Task~1 $\to$ Cell~1 and Task~2 $\to$ Cell~2.\newline
    \textbf{2.} Tier-0 AST hook detects shared symbol \texttt{sync\_do\_groupby}; Cell~2 invokes four read-only \texttt{peek\_peer\_cell} diff reads.\newline
    \textbf{3.} Cell~1 settles first; Cell~2 receives trunk-advance notice, unifies \texttt{(..., case\_sensitive=False, reverse=False)} in trial merge, and passes full \texttt{pytest} on main. \\
    \midrule
    \raggedright\textbf{PA150-H3-10} \newline (Outlines \texttt{types.guid} + read-only release brief) &
    \textbf{1.} External issue opens coding ticket for \texttt{types.guid}.\newline
    \textbf{2.} User asks in chat for a release cross-check brief: \emph{``read-only tracking---do not modify business code.''}\newline
    \textbf{3.} Top-level router conflates foreground caveat with active mandate, suppressing all file edits ($0$ files changed $\to$ \texttt{AttributeError}). &
    \textbf{1.} Cell~1 executes \texttt{types.guid} implementation inside isolated worktree.\newline
    \textbf{2.} MMU classifies foreground brief as derived read-only request and spawns zero-worktree Observer Cell ($\sigma_2 = \mathrm{obs}$) under $\mathcal{M}_{\mathrm{ro}}$.\newline
    \textbf{3.} Cell~1 lands $84/84$ passing tests undisturbed; Observer Cell reads settled milestone via Synapse and delivers verified brief. \\
    \midrule
    \raggedright\textbf{Micro-30 M-1} \newline (GPT-6-Luna: dual-app errands + chatter) &
    \textbf{1.} Router decodes full event log on every chatter message ($889.5$k tokens/episode).\newline
    \textbf{2.} Compact-model \texttt{MasterAssessor} drops exact recipient/date constraints while paraphrasing prose tickets or mis-rejects personal app requests ($0.0\%$ M-1 pass, $63.3\%$ hygiene). &
    \textbf{1.} Single-pass MMU drops bystander chatter and duplicate alerts at entry ($100.0\%$ hygiene).\newline
    \textbf{2.} Actionable errands route verbatim into independent Context Cells ($T_0^{(e)} \oplus [e_t]$) with zero ticket paraphrasing, sustaining $60.0\%$ M-1 pass with $53.7\%$ fewer tokens ($412.2$k). \\
    \bottomrule
  \end{tabular}
\end{table}

\end{document}